\documentclass[12pt,numbers]{article}

\usepackage[margin=2.5cm]{geometry}
\usepackage{braket, siunitx}
\usepackage{graphicx}
\usepackage{dcolumn}
\usepackage{bm}
\usepackage{amssymb, amsmath}
\usepackage{multirow}
\usepackage[acronym]{glossaries}
\usepackage[autostyle]{csquotes}
\usepackage{algorithm}
\usepackage{algpseudocode}
\usepackage{setspace}
\usepackage{authblk}
\usepackage{csquotes}
\usepackage{hyperref}
\usepackage{subfigure}
\usepackage{dirtytalk}
\usepackage[giveninits=true, backend=biber, date=year, style=numeric-comp, sorting=none, minbibnames=30, maxbibnames=30,isbn=false,url=false,eprint=false]{biblatex}
\renewbibmacro{in:}{}
\AtEveryBibitem{\clearlist{language}}
\AtEveryBibitem{\clearfield{note}}
\bibliography{references}
\usepackage[font=small]{caption}

\usepackage{doi}

\makeatletter
\def\@fnsymbol#1{\ensuremath{\ifcase#1\or * \or *\or 3\or 4\or 5\or 6\or *\or z\or \ddagger\or
\mathsection\or \mathparagraph\or \|\or **\or \dagger\dagger
\or \ddagger\ddagger \else\@ctrerr\fi}}
\makeatother

\DeclareMathOperator*{\argmin}{arg\,min}
\DeclareMathOperator*{\argmax}{arg\,max}

\newacronym{ace}{ACE}{Alternating Conditional Expectations}
\newacronym{ag}{AG}{Artificial Graphite}
\newacronym{alven}{ALVEN}{Algebraic Learning Via Elastic Net}
\newacronym{cc}{CC}{Constant Current}
\newacronym{cv}{CV}{Constant Voltage}
\newacronym{ct}{CT}{Computed Tomography}
\newacronym{dtw}{DTW}{Dynamic Time Warping}
\newacronym{dvf}{DVF}{Differential Voltage Fitting} 
\newacronym{eis}{EIS}{Electrochemical Impedance Spectroscopy}
\newacronym{en}{EN}{Elastic Net}
\newacronym{eol}{EoL}{End of Life}
\newacronym{hppc}{HPPC}{Hybrid Pulse Power Characterization}
\newacronym{ica}{ICA}{Incremental Capacity Analysis}
\newacronym{icet}{ICET}{Ion-Coupled Electron Transfer}
\newacronym{lcen}{LCEN}{Lasso-Clip-EN}
\newacronym{lhs}{LHS}{Latin Hypercube Sampling}
\newacronym{lib}{LiB}{Lithium-ion Battery}
\newacronym{lz}{LZ}{Lempel-Ziv}
\newacronym{mape}{MAPE}{Mean Absolute Percentage Error}
\newacronym{ml}{ML}{Machine Learning}
\newacronym{nmc}{NMC}{Nickel Manganese Cobalt}
\newacronym{pc}{PC}{Principal Component}
\newacronym{pls}{PLS}{Partial Least Squares}
\newacronym{rf}{RF}{Random Forest}
\newacronym{rmse}{RMSE}{Root Mean Square Error}
\newacronym{rpt}{RPT}{Reference Performance Test}
\newacronym{rr}{RR}{Ridge Regression}
\newacronym{se}{SE}{Standard-Error}
\newacronym{sei}{SEI}{Solid Electrolyte Interphase}
\newacronym{shap}{SHAP}{SHapley Additive exPlanations}
\newacronym{si}{SI}{Supplementary Information}
\newacronym{soc}{SoC}{State of Charge}
\newacronym{soh}{SoH}{State of Health}
\newacronym{spa}{SPA}{Smart Process Analytics}
\newacronym{spls}{SPLS}{Sparse Partial Least Squares}
\newacronym{svr}{SVR}{Support Vector Regression}
\newacronym{tof}{ToF}{Time-of-Flight}
\newacronym{vif}{VIF}{Variance Inflation Factor}
\newacronym{xgb}{XGB}{XGBoost}
\newacronym{esf}{ESF}{\textit{Extreme Slow Formation}}

\providecommand{\keywords}[1]
{
  \small	
  \textbf{\textit{Keywords---}} #1
}

\title{Systematic Feature Design for Cycle Life Prediction of Lithium-Ion Batteries During Formation}

\author[1]{\small Jinwook Rhyu}
\author[1,2]{\small Joachim Schaeffer}
\author[1]{\small Michael L. Li}
\author[3,4]{\small Xiao Cui}
\author[3,4]{\small William~C. Chueh}
\author[1,5]{\small Martin~Z.~Bazant}
\author[1]{\small Richard D. Braatz\thanks{Corresponding author: Department of Chemical Engineering, Massachusetts Institute of Technology, 77 Massachusetts Avenue, Room E19-551, Cambridge, MA 02139, \texttt{braatz@mit.edu}.}}

\affil[1]{\footnotesize Department of Chemical Engineering, Massachusetts Institute of Technology, Cambridge, MA 02139, USA.}
\affil[2]{\footnotesize Control and Cyber-Physical Systems Laboratory, Technical University of Darmstadt, 64289, Germany.}
\affil[3]{\footnotesize Department of Materials Science and Engineering, Stanford University, Stanford, CA 94305, USA.}
\affil[4]{\footnotesize Applied Energy Division, SLAC National Accelerator Laboratory, Menlo Park, CA 94025, USA.}
\affil[5]{\footnotesize Department of Mathematics, Massachusetts Institute of Technology, Cambridge, MA 02139, USA.}

\begin{document}
\maketitle

\renewcommand{\abstractname}{Summary}
\begin{abstract}
Optimization of the formation step in lithium-ion battery manufacturing is challenging due to limited physical understanding of solid electrolyte interphase formation and the long testing time ($\sim$100 days) for cells to reach the end of life.
We propose a systematic feature design framework that requires minimal domain knowledge for accurate cycle life prediction during formation.
Two simple $Q(V)$ features designed from our framework, extracted from formation data without any additional diagnostic cycles, achieved a median of $9.20\%$ error for cycle life prediction, outperforming thousands of autoML models using pre-defined features.
We attribute the strong performance of our designed features to their physical origins -- the voltage ranges identified by our framework capture the effects of formation temperature and microscopic particle resistance heterogeneity.
By designing highly interpretable features, our approach can accelerate formation research, leveraging the interplay between data-driven feature design and mechanistic understanding.
\end{abstract}

\keywords{Feature design, Feature engineering, Interpretable machine learning, Systematic framework, Cycle life prediction, Lifetime prediction, Prediction, SEI formation, Lithium-ion batteries, Fused lasso}

\newpage
\section{Introduction}\label{sec:introduction}

Accurate lifetime prediction of lithium-ion batteries accelerates battery optimization and improves safety \cite{bandhauer_critical_2011, deng_li-ion_2015, deng_safety_2018, attia_closed-loop_2020}.
Although this task is challenging due to complicated and convolved degradation mechanisms, various studies have demonstrated the potential in using data-driven approaches \cite{davies_state_2017, severson_data-driven_2019, sulzer_promise_2021, zhang_acoustic_2021, greenbank_automated_2022, jones_impedance-based_2022, paulson_feature_2022,schaeffer_cycle_2024, schaeffer_interpretation_2024}, physics-based approaches \cite{ning_cycle_2004, deshpande_battery_2012, reniers_review_2019, downey_physics-based_2019, el-dalahmeh_physics-based_2023}, and hybrid approaches \cite{liu_data-model-fusion_2012, liao_review_2014, zheng_integrated_2015, chang_new_2017, aitio_combining_2020, sulzer_challenge_2021, aykol_perspectivecombining_2021, schaeffer_lithium-ion_2024}.
For accurate battery health monitoring, diagnostic techniques such as \gls{dvf} \cite{bloom_differential_2005, dahn_user-friendly_2012, weng_differential_2023, lin_identifiability_2024}, \gls{ica} \cite{dubarry_synthesize_2012, dubarry_best_2022}, \gls{eis} \cite{jones_impedance-based_2022, huang_review_2007, chang_electrochemical_2010, schaeffer_machine_2023}, and \gls{hppc} \cite{dees_electrochemical_2008, holland_experimental_2002} were developed for physics-based feature extraction during battery operation.
Further optimization of these diagnostic techniques includes novel \gls{soh} feature development \cite{che_battery_2023, huang_online_2017, tian_state--health_2020, son_integrated_2022} and diagnostic time reduction \cite{sun_optimization_2023, rhyu_optimum_2024}.

Compared to the extensive research on lifetime prediction during operation, there have been few studies on lifetime prediction during the manufacturing process (i.e., \textit{extreme} early cycle life prediction) because of the limited availability of public manufacturing data.
In fact, the cycle life can vary greatly based on the protocol used during formation, in which a passivation layer of \gls{sei} is rapidly formed on the anode to limit further degradation during use.
For example, Weng et al.~\cite{weng_predicting_2021} showed that the \gls{nmc}/graphite pouch cells with the fast formation protocol proposed by Wood et al.~\cite{wood_formation_2019, an_fast_2017} had in average $\sim$25$\%$ longer cycle lives than the pouch cells with a baseline formation protocol when aging the cells in both room temperature and high-temperature (45$^{\circ}\text{C}$) cases.
Recently, Cui et al.~\cite{cui_data-driven_2024} showed that the cycle life can vary two-fold by only manipulating the formation protocols.
Moreover, formation is the most expensive step in cell manufacturing in terms of cost, time, and energy  \cite{liu_current_2021}, thus emphasizing the importance of optimizing formation.

Extreme early cycle life prediction using machine learning can accelerate the optimization of formation protocols and help build mechanistic understanding, to the extent that features correlated with cycle life can be identified in the formation data. Weng et al.~\cite{weng_predicting_2021} proposed the low \gls{soc} resistance ($R_{\text{LS}}$) as a feature for extreme early cycle life prediction, achieving $\sim$8\% error\footnote{Note that this small value was obtained by having the cells from the same formation protocol both at \say{validation} set and \say{train/test} set in Ref.~\cite{weng_predicting_2021}.} ($\sim$15\% error for dummy model) over the dataset composed of 40 cells with two different formation protocols using the single feature based on the strong linear correlation between $R_{\text{LS}}$ and the cycle life.
$R_{\text{LS}}$ is not only predictive but also practical since it does not require additional equipment to be installed in the manufacturing process, unlike other characterization techniques such as \gls{eis} \cite{an_fast_2017, schranzhofer_electrochemical_2006, zhang_identifying_2020}, X-ray tomography \cite{pietsch_x-ray_2017, wood_x-ray_2018, condon_dataset_2024}, or acoustic \gls{tof} \cite{hsieh_electrochemical-acoustic_2015, ke_potential_2022}.
Furthermore, $R_{\text{LS}}$ is easy to interpret since the feature is sensitive to lithium loss during formation \cite{weng_predicting_2021}.

While $R_{\text{LS}}$ is an interpretable, predictive, and easy-to-implement feature, there are areas for further improvement.
For example, the $R_{\text{LS}}$ feature is highly sensitive to the \gls{soc} value where the resistance was measured and thus its use requires accurate \gls{soc} estimation. 
According to Weng et al.~\cite{weng_predicting_2021}, the $R_{\text{LS}}$ difference between the two formation protocols is roughly $\SI{10}{m\Omega}$ (see Figure 3c in Ref.~\cite{weng_predicting_2021}).
Given that the resistance varies roughly $\SI{75}{m\Omega}$ with respect to $4\%$ change in \gls{soc} at low-\gls{soc} region (see Figure S9a in Ref.~\cite{weng_predicting_2021}), even $0.5\%$ error in \gls{soc} estimation can smear out the strong negative correlation between $R_{\text{LS}}$ and the cycle life.
Therefore, additional low-rate cycle, which roughly takes a day to complete, is needed after the formation step to accurately calculate the full-cell \gls{soc}.
Furthermore, $R_{\text{LS}}$ only works for cells that undergo the same formation temperature (see \gls{si} \ref{appdx:R_LS}).
This fact indicates that a single $R_{\text{LS}}$ feature is insufficient for optimizing formation, especially given that the formation temperature greatly impacts the cycle life \cite{cui_data-driven_2024}.
These limitations motivate the development of new features that are (1) obtainable without additional diagnostic cycles and (2) capable of comparing formation protocols with different temperatures while still retaining the strengths of $R_{\text{LS}}$.

In this work, we propose a systematic framework to automatically \textit{design} predictive yet interpretable features (i.e., transformation of input data) for regression problems.
This framework is especially useful for investigating systems with complicated physics, such as \gls{sei} formation \cite{weng_modeling_2023}, where automatic feature extraction can be more effective than hand-crafted features that are limited by the many unknown aspects of the underlying physics.
The performance of the designed features is then compared with agnostic and autoML approaches.
Finally, we conduct a physics-based investigation using a distributed resistance model to explain the outstanding performance of the designed features.

\section{Methods}\label{sec:methods}
\subsection{Description on the dataset}\label{sec:methods_data_description}

\begin{figure}[h!]
\includegraphics[width=0.5\textwidth]{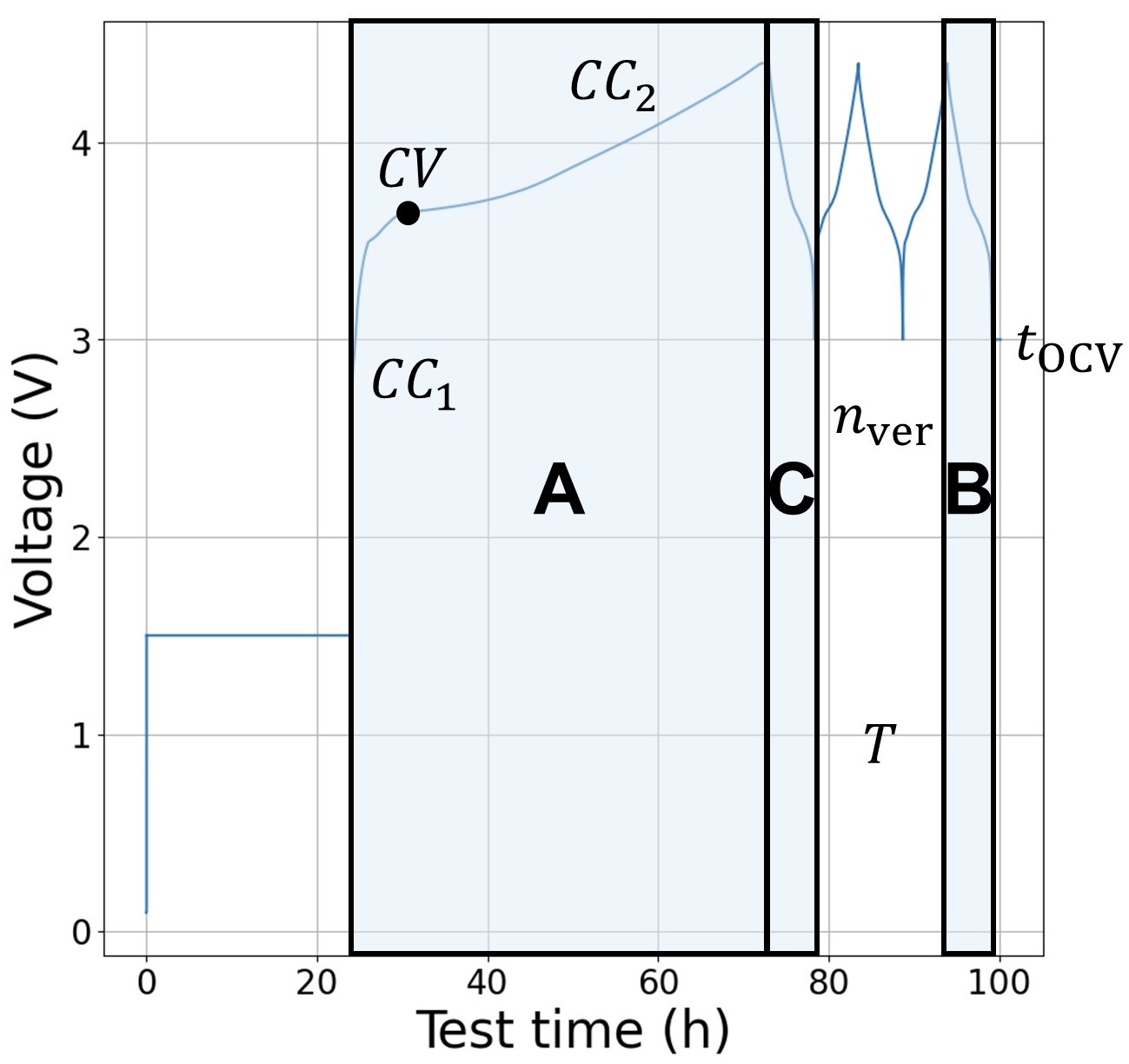}
\centering
\caption{Schematic of formation protocols used for generating the dataset, modified from Figure 1 in Ref.~\cite{cui_data-driven_2024}.
Total of 62 protocols were tested by manipulating the six formation protocol parameters: C-rate for the two-step charging ($CC_1$ and $CC_2$), the cutoff voltage between the two CC steps ($CV$), number of cycles between the first charge and last discharge step ($n_{\text{ver}}$), formation temperature ($T$), and the rest time ($t_{\text{OCV}}$).
Three common steps among various formation protocols are indicated in blue: the first charge step (Step A), the last discharge step (Step B), and the first discharge step (Step C).
}
\label{fig:dataset}
\end{figure}

A dataset of 186 single crystal $\mathrm{Li[Ni_{0.5}Mn_{0.3}Co_{0.2}]O_2}$(SC-NMC532)/\gls{ag} pouch cells with 62 different formation protocols and the identical aging protocol was used in this study \cite{cui_data-driven_2024}.
Among 186 pouch cells, we used 178 cells that reached the end of life (i.e., having a discharge capacity below 80\% of its initial value measured at 0.75C \gls{cc} discharge step). 
Six parameters were varied in the dataset: current for two-step \gls{cc} at the first charge step ($CC_1$ and $CC_2$), the cutoff voltage between the two \gls{cc} steps ($CV$), number of cycles between the first charge and last discharge step ($n_{\text{ver}}$), temperature during the formation step ($T$), and the rest time after the formation step ($t_{\text{OCV}}$).
The six parameters were chosen using \gls{lhs} for its ability to explore the parameter space efficiently \cite{stein_large_1987}.
Detailed data interrogation results using the six formation protocol parameters and cycle life can be found in the \gls{si} \ref{appdx:data_interrogation}.
We define three common steps that appear in all 62 formation protocols (see Figure \ref{fig:dataset}): the first charge step (Step A), the last discharge step (Step B), and the first discharge step (Step C).
All variations among the formation protocols are encoded in Step A whereas Steps B and C undergo the identical operating protocol.

As a preliminary analysis, we constructed physics-\textit{agnostic} \gls{ml} models that map the six formation protocol parameters to the cycle life.
While the agnostic models cannot capture cell-to-cell variability, they serve as a good baseline for evaluating model performance since the features directly encode the formation protocol.
For the agnostic approach, we used nested cross-validation to avoid information leakage\footnote{From the same dataset being used for optimizing hyperparameters and evaluating the model performance.} \cite{kapoor_leakage_2023, gibney_could_2022, geslin_selecting_2023}. 

A total of 62 formation protocols were grouped into five sets, where one set was used as the test set and the remaining four as the training set in each outer loop to evaluate the model performance. These splits are shown in \gls{si} Table \ref{tab:split_formation_protocols}.
Within each outer loop, the inner loop was performed by dividing the training set into ten subsets, with one being used as the validation set and the others as the training set in each inner loop for optimizing the hyperparameters, as shown in Table \ref{tab:hyperparams}. 
A total of 52 agnostic models were constructed based on the combination of categories listed in Table \ref{tab:agnostic_options}.
\gls{spa} software \cite{sun_smart_2021} was used for model construction, which which includes a suite of static \gls{ml} algorithms: \gls{rr} \cite{hoerl_ridge_1970}, \gls{en} \cite{zou_regularization_2005}, \gls{pls} \cite{geladi_partial_1986}, \gls{spls} \cite{chun_sparse_2010}, \gls{rf} \cite{breiman_random_2001}, \gls{svr} \cite{vapnik_nature_2000}, \gls{xgb} \cite{chen_xgboost_2016}, \gls{alven} \cite{sun_alven_2020}, and \gls{lcen} \cite{seber_lcen_2024}.

\begin{table}[h!]
\begin{center}
\footnotesize
\begin{tabular}{|c|c|c|}
\hline
ML algorithm & Hyperparameters & \#\\ 
 \hline
\gls{rr} & L2 penalty term & 1 \\
\gls{en} & L1 penalty term, L2 penalty term & 2 \\
\gls{pls} & Number of \glspl{pc} & 1 \\
\gls{spls} & Number of \glspl{pc}, Sparsity & 2 \\
\gls{rf} & Maximum depth of trees & 1 \\
\gls{svr} & Regularization parameter, Kernel coefficient, Margin of tolerance & 3 \\
\gls{xgb} & Maximum depth of trees, L2 penalty term & 2 \\
\gls{alven} & Degree, L1 penalty term, L2 penalty term & 3 \\
\gls{lcen} & Degree, L1 penalty term, L2 penalty term & 3 \\
 \hline
\end{tabular}
\caption{Hyperparameters of \gls{ml} algorithms.}
\label{tab:hyperparams}
\end{center}
\end{table}

\begin{table}[h!]
\begin{center}
\footnotesize
\begin{tabular}{|c|c|c|}
\hline
Category Description & Options & \#\\ 
 \hline
Formation protocol parameters & \begin{tabular}{@{}c@{}}Full set ($CC_1$, $CC_2$, $CV$, $n_{\text{ver}}$ $T$, $t_{\text{OCV}}$),\\ Subset ($CC_1$, $CC_2$, $CV$, $T$)\end{tabular} & 2 \\ \hline
Log-transformed output & Yes, No & 2 \\ \hline
\gls{ml} algorithms & \begin{tabular}{@{}c@{}c@{}} (Linear) \gls{rr}, \gls{en}, \gls{pls}, \gls{spls},\\ (Nonlinear) \gls{rf}, \gls{svr}, \gls{xgb},\\ (Nonlinear quantifiable) \gls{alven}, \gls{lcen} with degree of 1, 2, 3\end{tabular} & 13 \\
 \hline
\end{tabular}
\caption{Description of three categories for constructing agnostic models. The subset of formation protocol parameters was selected based on \gls{shap} (SHapley Additive exPlanations) analysis \cite{lundberg_unified_2017} conducted by Cui et al.~\cite{cui_data-driven_2024}.}
\label{tab:agnostic_options}
\end{center}
\end{table}

In this work, the \textit{best} model is defined as the one with the smallest summation of the median and maximum \gls{mape} among the five outer loops to consider both average and extrapolation performances.
The best agnostic model had median and maximum \gls{mape} of 11.06 and 11.35, respectively.
Although the best agnostic model performs well, the model cannot be used for diagnosing individual cell degradation as it cannot capture the cell-to-cell variation.
Furthermore, the agnostic model cannot be used for formation protocols that do not follow the specific template used by Cui et al.~\cite{cui_data-driven_2024}, which limits its usage in optimizing the formation process.

\subsection{Systematic Feature Design Framework}\label{sec:methods_feature_design}
\begin{figure}[h!]
\includegraphics[width=\textwidth]{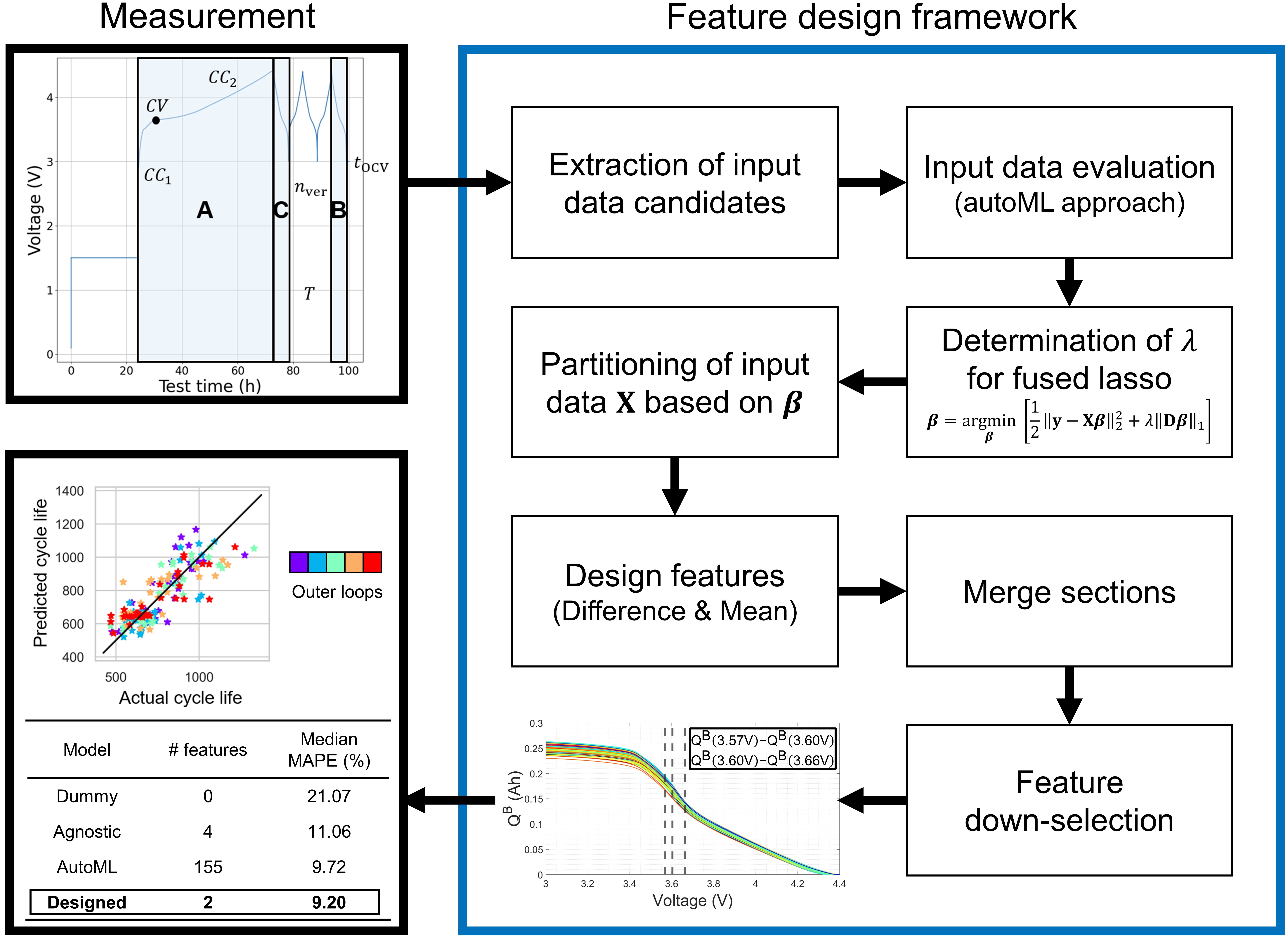}
\centering
\vspace{-0.5cm}
\caption{Systematic feature design framework for extreme early cycle life prediction.}
\label{fig:framework}
\end{figure}

This section describes the workflow of the proposed systematic feature design framework as displayed in Figure \ref{fig:framework}.
We begin with extracting the input data (i.e., the source for designing features) candidates from the raw measurements.
Then, the promisingness of each candidate is evaluated based on the autoML approach.
After determining the promising input data types, we determine which value of penalty term $\lambda$ to use for the fused lasso model.
Then, the features are designed based on the fused lasso coefficient $\boldsymbol{\beta}$ that maps the selected input ($Q(V)$ in Step B) to the output (cycle life).
Lastly, the features are down-selected to finalize the designed features.

\subsubsection{Extraction of input data}\label{sec:methods_extraction}

The dataset contains seven measurements as a function of time ($t$) during the formation process: current ($I$), voltage ($V$), capacity ($Q$), energy ($E$), temperature ($T$), cycle index, and step index, where each measurement was taken at every $\SI{3}{mV}$ or 5 seconds for \gls{cc} step and at every $\SI{3}{mA}$ or 5 seconds for \gls{cv} step, whichever comes first.
Among various possible input data combinations, where the input variable (i.e., $x$ in $f(x)$) and the function (i.e., $f$ in $f(x)$) are selected from the measurements mentioned above, domain knowledge can be used to narrow down the most promising input data candidates.
For example, the current $I$ can be discarded because $I$ is constant for most of the process.
While $t$, $V$, and $Q$ are monotonic functions within each Step A, B, and C, and thus can be considered as the input variable, the range for $V$ is identical for all formation protocols whereas the range for $t$ and $Q$ within each step may vary from cell to cell. 
Therefore, the $t$ and $Q$ should be normalized when used as the input variable (i.e., $\tilde{t}, \tilde{Q} \in [0,1]$).
However, we may not use $\tilde{t}$ and $\tilde{Q}$ as input variables in Step A since single $\tilde{t}$ cannot specify the \gls{soc} of the cell whereas single $\tilde{Q}$ masks the impact of C-rate, which greatly affects the electrode utilization range \cite{cui_data-driven_2024}, given that the C-rates vary by order of magnitudes in Step A.
In addition, any input data of $f(\tilde{t})$ and $f(\tilde{Q})$ are redundant for the constant current steps due to $Q=It$, implying that only one may be considered for Steps B and C.
Thus, six input data types\footnote{Each input data candidate was processed using interpolation at 1,000 uniformly distributed points along the input variable ($x$).} are considered for the candidates: $Q^{\text{A}}(V)$, $t^{\text{A}}(V)$, $Q^{\text{B}}(V)$, $V^{\text{B}}(\tilde{t})$, $Q^{\text{C}}(V)$, and $V^{\text{C}}(\tilde{t})$.

\subsubsection{Input data evaluation}\label{sec:methods_inputdata}
To design regression models with high prediction accuracy and robust generalization performance, we propose a method to select the input data that is \textit{promising} for designing highly predictive features in this section.
First, we construct \textit{autoML} models using each input data candidate to evaluate their overall performance.
Here, the autoML model refers to the ML model using features that are automatically extracted and selected without human assistance~\cite{he_automl_2021}. 
The $\textsf{tsfresh}$ package \cite{christ_time_2018}, which is a highly parallelized package for extracting roughly $800$ features from time series data and guiding feature selection based on the feature importance \cite{wang_towards_2022, zhang_unsupervised_2020, liu_sensor_2020}, was used in this step.
From the features generated using $\textsf{tsfresh}$, we can construct a total of 2,448 autoML models based on the five categories listed in Table \ref{tab:autoML_options}.

\begin{table}[h!]
\begin{center}
\footnotesize
\renewcommand{\arraystretch}{1.2}
\begin{tabular}{|c|c|c|}
\hline
Category Description & Options & \#\\ 
 \hline
Input data types & $Q^{\text{A}}(V)$, $t^{\text{A}}(V)$, $Q^{\text{B}}(V)$, $V^{\text{B}}(\tilde{t})$, $Q^{\text{C}}(V)$, $V^{\text{C}}(\tilde{t})$ & 6 \\
p-values from univariate statistical test & $10^{0}$, $10^{-0.5}$, $\cdots$, $10^{-7.5}$, $10^{-8}$ & 17 \\
Further feature selection using $\textsf{tsfresh}$ & Yes, No & 2 \\
Log-transformed output & Yes, No & 2 \\
ML algorithms & \gls{en}, \gls{rf}, \gls{svr}, \gls{xgb}, \gls{alven}, \gls{lcen} & 6 \\
 \hline
\end{tabular}
\renewcommand{\arraystretch}{1}
\caption{Description of five categories for constructing autoML models. The features extracted from the \textsf{tsfresh} package are first pre-screened based on the univariate statistical test (F-statistics) with the p-value threshold chosen as in the second category. The features are further selected using $\textsf{select\_features}$ function in \textsf{tsfresh} package if the third category is yes.}
\label{tab:autoML_options}
\end{center}
\end{table}

\begin{table}[h!]
\begin{center}
\footnotesize
\renewcommand{\arraystretch}{1.2}
\begin{tabular}{|c|c|c|c|c|c|}
\hline
Models & \# features & Median RMSE & Max RMSE & Median MAPE & Max MAPE\\ \hline
Agnostic & 4 & 107.10 & 124.81 & 11.06 & 11.35\\ \hline
AutoML -- $Q^{\text{A}}(V)$ & 128 & 131.32 & 162.55 & 12.39 & 13.74\\ 
AutoML -- $t^{\text{A}}(V)$ & 160 & 112.47 & 183.07 & 13.00 & 14.68\\
AutoML -- $Q^{\text{B}}(V)$ & 155 & \textit{\textbf{86.17}} & 136.92 & \textit{\textbf{9.72}} & \textbf{\textit{10.85}}\\ 
AutoML -- $V^{\text{B}}(\tilde{t})$ & 190 & \textit{99.87} & 127.70 & \textit{9.87} & 11.62\\ 
AutoML -- $Q^{\text{C}}(V)$ & 118 & \textit{95.81} & 132.05 & \textit{10.39} & \textit{11.11}\\ 
AutoML -- $V^{\text{C}}(\tilde{t})$ & 281 & \textit{92.23} & 136.41 & \textit{10.12} & \textit{11.27}\\ \hline
\end{tabular}
\renewcommand{\arraystretch}{1}
\caption{The predictive performance of the best models among 52 agnostic models and 408 autoML models for each input data candidates: $Q^{\text{A}}(V)$, $t^{\text{A}}(V)$, $Q^{\text{B}}(V)$, $V^{\text{B}}(\tilde{t})$, $Q^{\text{C}}(V)$, and $V^{\text{C}}(\tilde{t})$. Italicized numbers indicate values smaller than the best agnostic model (i.e., the first row), and bold numbers indicate the smallest in each column.}
\label{tab:promisingness_inputdata}
\end{center}
\end{table}

Table \ref{tab:promisingness_inputdata} displays the number of features and four performance metrics (median and maximum \gls{rmse} and \gls{mape} among five outer loops) of the best agnostic model and the best autoML model for each input data candidate (see Figure \ref{fig:histogram_autoML} for histogram of four performance metrics over all 2,448 autoML models).
Given that we are constructing autoML models over 408 combinations of categories listed in Table \ref{tab:autoML_options}, where the number of features spans from one to more than hundreds, the input data type is not considered promising if the best autoML model does not outperform the best agnostic model in any of the four performance metrics.

The best autoML model with four input data types from Steps B and C outperforms the best agnostic model in at least two performance metrics whereas the best model with two input data types from Step A is strictly worse (Table \ref{tab:promisingness_inputdata}).
While this may seem counter-intuitive as the protocols only vary in Step A, it can be explained by the fact that the features extracted from Step A encode the operational variation whereas the features from Steps B and C encode conditional variation (i.e., the cell's response to the formation protocol) that capture implicit information on both protocol-to-protocol and cell-to-cell variability \cite{geslin_selecting_2023}.
As a result, autoML models from Steps B and C are more predictive than the autoML models from Step A since the latter heavily depend on the trained formation protocols.
While there are four input data types from Steps B and C, we focus on $Q^{\text{B}}(V)$ in the subsequent parts since it outperforms the other candidates (see \gls{si} \ref{appdx:feature_design_otherinput} for feature design results with $V^{\text{B}}(\tilde{t})$, $Q^{\text{C}}(V)$, and $V^{\text{C}}(\tilde{t})$).

\subsubsection{Determination of linear regression coefficient \texorpdfstring{$\boldsymbol{\beta}$}{TEXT} from input data}\label{sec:methods_beta}

\begin{figure}[h!]
\includegraphics[width=\textwidth]{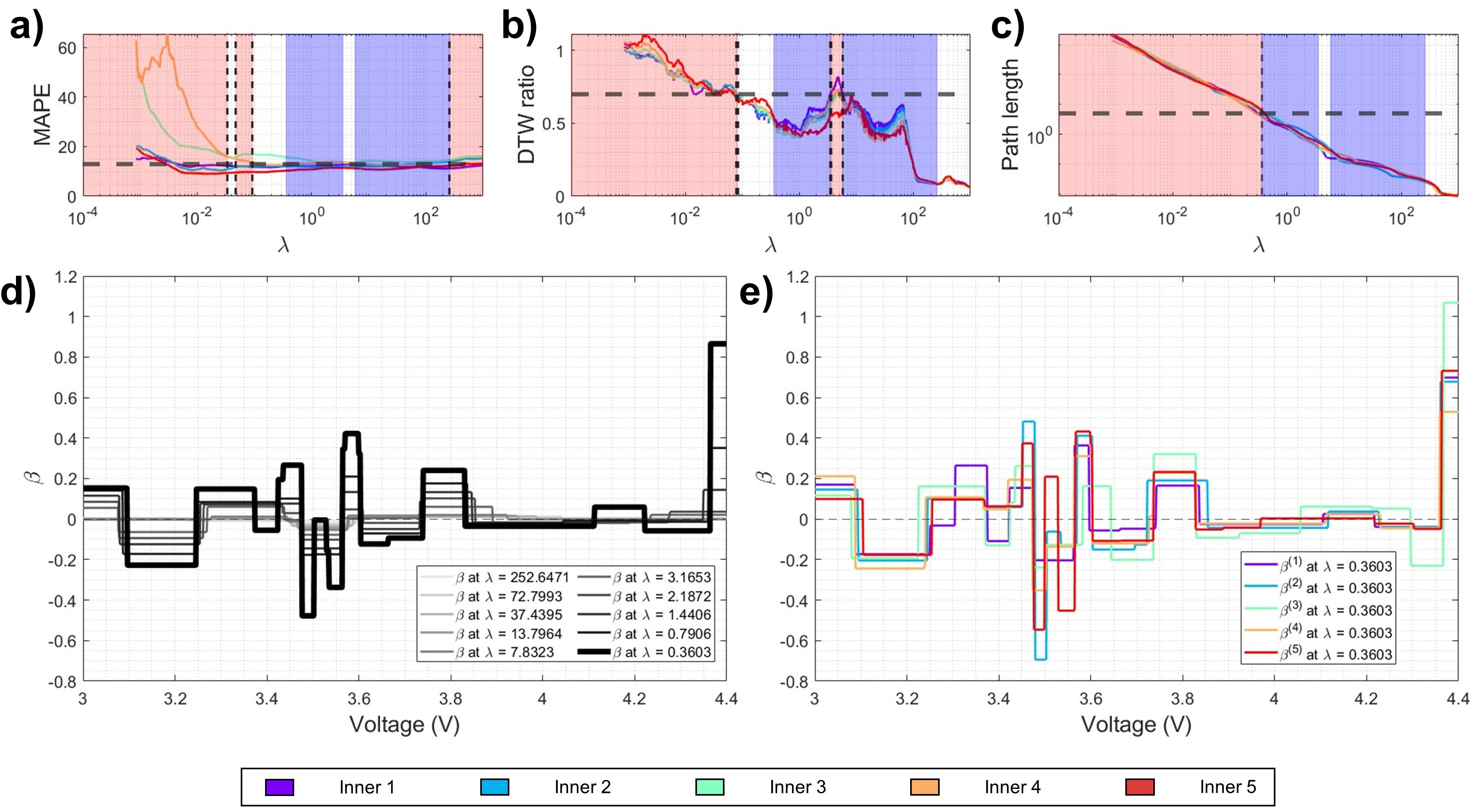}
\centering
\caption{Determination of $\lambda$ for Outer loop 1 with $Q^{\text{B}}(V)$ based on Table \ref{tab:downselect_pairs}. (a) Predictiveness, (b) Robustness, and (c) Interpretability metrics as a function of $\lambda$. The horizontal dotted line indicates the constraint and the vertical dotted lines are the indices where the constraint is activated. The red region is where the constraint is violated and the blue region is where all three constraints are satisfied. (d) Demonstration of $\boldsymbol{\beta}$ at various $\lambda$ values in the blue region. (e) Demonstration of $\boldsymbol{\beta}^{(k)}$'s at smallest $\lambda$ from the blue region for robustness. Colors in panels (a)--(c) and (e) indicate different inner loops.}
\label{fig:three_metrics}
\end{figure}

As an alternative to using nonlinear features, linear regression on high-dimensional data can also learn a nonlinear response \cite{schaeffer_method_2024}.
One advantage of using linear regression models is that we obtain a regression coefficient $\boldsymbol{\beta}$ that can give insights on how each portion of the input data contributes to the output estimation \cite{schaeffer_interpretation_2024,schaeffer_latent_2022}.
Among various linear models, we use the generalized lasso with penalty on the differences of adjacent coefficients (often referred to as {\em fused lasso}~\cite{tibshirani_solution_2011}),
\renewcommand{\arraystretch}{0.7}
\begin{equation} \label{eq:genlasso}
\min_{\boldsymbol{\beta} \in \mathbb{R}^p} \frac{1}{2}||\mathbf{y} - \mathbf{X}\boldsymbol{\beta}||_2^2 + \lambda||\mathbf{D}\boldsymbol{\beta}||_1 \quad \text{with } \mathbf{D} = \!\begin{bmatrix}
        -1 & 1 & 0 & \cdots & 0 \\
        0 & -1 & 1 & \ddots  & \vdots \\
        \vdots & \ddots & \ddots &\ddots &0 \\
        0 & \cdots & 0 & -1 & 1 \\
    \end{bmatrix} \!\in \mathbb{R}^{(p-1) \times p},
\end{equation}
to obtain regression coefficients $\boldsymbol{\beta}$. 
Standardized $\mathbf{X}$ and $\mathbf{y}$ are used when solving Equation \ref{eq:genlasso}, where every column in $\mathbf{X}$ is divided by the maximum column-wise standard deviation (i.e., $\max_{j=1, \cdots{}, p} \text{std}(\mathbf{X}_{:,j})$).
Unlike standardizing each column of the input data, this method preserves the unique characteristic (e.g., trend of column-wise variance) of the raw data.
This model yields sparsity in regression coefficient difference (i.e., piecewise constant regression coefficients) roughly in line with physical expectations that neighboring regression coefficients should be similar and only change at specific locations. 
In turn, the fused lasso regression coefficients can guide the partitioning of the input data into smaller sections and improve interpretability. \renewcommand{\arraystretch}{1}
Even from the same $\mathbf{X} \in \mathbb{R}^{n \times p}$ and $\mathbf{y} \in \mathbb{R}^{n \times 1}$, we can obtain various $\boldsymbol{\beta}$ by changing the penalty term $\lambda$.
For example, having a larger $\lambda$ would place a stronger penalty for $\boldsymbol{\beta}$ changing its values, making $\boldsymbol{\beta}$ \textit{simpler} (see Figure \ref{fig:three_metrics}d).

Considering the $\lambda$ dependency of $\boldsymbol{\beta}$, we should determine which value of $\lambda$ leads to $\boldsymbol{\beta}$ for guiding the design of predictive yet interpretable features.
For this process, we propose to determine $\lambda$ based on three criteria: predictiveness, robustness, and interpretability.
For predictiveness, the average of \gls{mape} among five inner loops is used as the representative metrics to evaluate how well the model predicts the cycle life of the cells with untrained formation protocols.
The \gls{dtw} distance ratio metric \cite{muller_information_2007, wang_alignment_1997} is used to quantify the robustness (i.e., how consistent the shape of $\boldsymbol{\beta}$ is for different training-test splits).
Lastly, the average of path length along $\boldsymbol{\beta}$ among five inner loops is used for assessing the interpretability of $\boldsymbol{\beta}$.
Details for each metric can be found in Table \ref{tab:downselect_pairs}.

\begin{table}[h!]
\begin{center}
\begin{tabular}{|c|c|c|}
\hline
Criteria & Metric description & Constraint\\ 
 \hline
Predictiveness & Average of \gls{mape} among five inner loops & 1SE rule \\ 
Robustness &  $\max_{k=1,\cdots{},5} \frac{\text{\gls{dtw} distance}(\boldsymbol{\beta}^{(k)}, \text{mean}(\boldsymbol{\beta}^{(-k)}))}{\text{\gls{dtw} distance}(\boldsymbol{0}, \text{mean}(\boldsymbol{\beta}^{(-k)}))}$  & $< 0.7$ \\
Interpretability & Average of path length along $\boldsymbol{\beta}^{(k)}$ & $< 5$ \\ \hline
\end{tabular}
\caption{Description of three metrics used for determining $\lambda$. $\boldsymbol{\beta}^{(k)}$ indicates the fused lasso coefficient learned from the training set in Inner loop $k \in \{1,\cdots{},5\}$.
DTW distance($\mathbf{a}$,$\mathbf{b}$) indicates the dynamic time warping distance between the vectors $\mathbf{a}$ and $\mathbf{b}$.
$\text{mean}(\boldsymbol{\beta}^{(-k)})$ denotes the average of $\boldsymbol{\beta}^{(i)}$'s where $i \in \{1,\cdots{},5\} \setminus \{k\}$. 
Path length of vector $\mathbf{a}\in \mathbb{R}^{n \times 1}$ is calculated by summing $|a_j - a_{j+1}|$ for $j \in \{1,\cdots{},n-1\}$.
1SE rule stands for one-standard-error rule, meaning that we are considering all $\lambda$ values that result in a mean \gls{mape} smaller than $\min_{\lambda}$(mean \gls{mape}) + standard error at $\lambda=\argmin$(mean \gls{mape}). The constraints for robustness and interpretability were chosen based on trial and error to ensure that $\boldsymbol{\beta}$ from any $\lambda$ in the blue region possess each characteristic.}
\label{tab:downselect_pairs}
\end{center}
\end{table}

Figures \ref{fig:three_metrics}a--c displays how each metric changes as a function of $\lambda$ in Outer loop 1 when using $Q^{\text{B}}(V)$ as the input data. 
The horizontal dotted line in each panel indicates the threshold of each metric as shown in Table \ref{tab:downselect_pairs}.
The red regions in each panel indicate that such $\lambda$ leads to the violation of the corresponding metric, whereas the blue regions indicate the $\lambda$ that satisfies all three constraints.
Figure \ref{fig:three_metrics}d demonstrates how the shape of $\boldsymbol{\beta}$ changes as a function of $\lambda$.
Among various $\lambda$ values that exist in the blue region, we observe that even $\boldsymbol{\beta}$ from the smallest $\lambda$ possesses high robustness (Figure \ref{fig:three_metrics}e) and interpretability due to conservative constraints as set in Table \ref{tab:downselect_pairs}.
Since $\boldsymbol{\beta}$ obtained from smaller $\lambda$ has a higher chance of capturing index-specific information, we use the smallest $\lambda$ from the blue region for obtaining $\boldsymbol{\beta}$ that will be used in the next step.

\subsubsection{Feature design based on \texorpdfstring{$\boldsymbol{\beta}$}{TEXT}}\label{sec:feature_design}

\begin{figure}[h!]
\includegraphics[width=\textwidth]{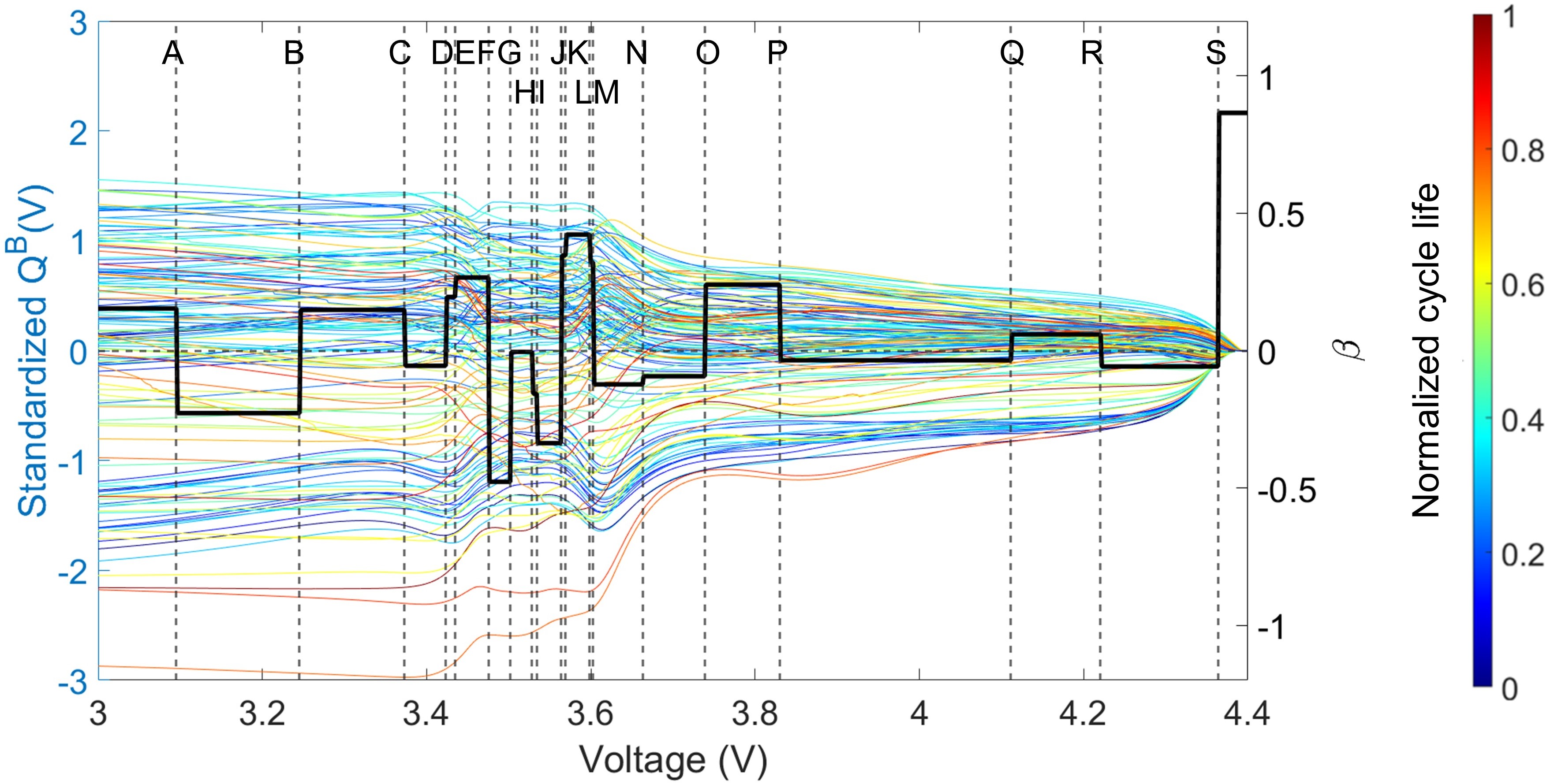}
\centering
\vspace{-0.7cm}
\caption{Visualization of $\tilde{Q}^{\text{B}}_i(V)$ (colored solid lines) for all cells in the training set, $\boldsymbol{\beta}$ (solid black line), and partition boundaries (vertical dotted lines) for Outer loop 1 with $Q^{\text{B}}(V)$ at $\lambda = 0.4018$. The color of $\tilde{Q}^{\text{B}}_i(V)$ curve indicates the normalized cycle life of the $i$th cell where 1 (red) is for the longest and 0 (blue) is for the shortest in the training set.}
\label{fig:partitioning}
\end{figure}

In this step, we use the determined $\boldsymbol{\beta}$ as a template for designing predictive and interpretable features.
The key advantage of using a linear model is that the regression problem can be split into smaller problems with a much simpler shape of $\mathbf{X}$ or $\boldsymbol{\beta}$ (i.e., $\hat{y} = \mathbf{X}\boldsymbol{\beta} = \sum_{m=1}^{M} \mathbf{X}_m\boldsymbol{\beta}_m$ where $\mathbf{X} = [\mathbf{X}_1, \cdots{}, \mathbf{X}_M]$ and $\boldsymbol{\beta} = [\boldsymbol{\beta}_1^{\mathsf{T}}, \cdots{}, \boldsymbol{\beta}_M^{\mathsf{T}}]^{\mathsf{T}}$).
Taking advantage of linearity, we can partition the input data into smaller sections based on the shape of $\boldsymbol{\beta}$, which makes $\boldsymbol{\beta}$ to be a flat line within each section.
Figure \ref{fig:partitioning} displays the standardized $Q^{\text{B}}(V)$ (Equation \ref{eq:standard}) of cells in the training set in colored solid lines and the determined $\boldsymbol{\beta}$ in black solid line for Outer loop 1. The standardized $Q^{\text{B}}(V)$ of $i$th cell is calculated as
\begin{equation} \label{eq:standard}
\tilde{Q}^{\text{B}}_i(V_j) = \frac{Q^{\text{B}}_i(V_j) - \bar{Q}^{\text{B}}(V_j)}{\max \text{std}(Q^{\text{B}}(V))},
\end{equation}
where $\bar{Q}^{\text{B}}$ is the column-wise average and $\max \text{std}(Q^{\text{B}}(V))$ is the maximum column-wise standard deviation of $Q^{\text{B}}(V)$ in the training set.
The vertical dotted lines in Figure \ref{fig:partitioning} indicate the boundaries for partitioning input data based on the indices where a jump (i.e., $|\boldsymbol{\beta}_{j+1} - \boldsymbol{\beta}_{j}| \geq 0.001 \times (\max\boldsymbol{\beta} - \min\boldsymbol{\beta})$) occurred.

Within each section partitioned by jumps in $\boldsymbol{\beta}$, it can be observed from Figure \ref{fig:partitioning} that $\tilde{Q}^{\text{B}}_i(V)$ can be approximated into a line (e.g., $\tilde{Q}^{\text{B}}_i(V) = a_iV + b_i$).
This linear approximation leads to the fact that only two features, the difference and the mean feature, are sufficient to represent each section.
For example, such approximation leads to the relationship:
\begin{equation} \label{eq:linear_approximation}
    \begin{split}
        \hat{y}_i^{\text{section}}(V_1, V_2) & := \sum_{j=\mathsf{ind}(V_1)}^{\mathsf{ind}(V_2)} \!\tilde{Q}^{\text{B}}_i(V_j) \boldsymbol{\beta}_j\ \approx \sum_{j=\mathsf{ind}(V_1)}^{\mathsf{ind}(V_2)} (a_i V_j + b_i) \boldsymbol{\beta}_j \\
        & = a_i\! \left(\sum_{j=\mathsf{ind}(V_1)}^{\mathsf{ind}(V_2)} V_j \boldsymbol{\beta}_j\right)\! + b_i\! \left(\sum_{j=\mathsf{ind}(V_1)}^{\mathsf{ind}(V_2)} \boldsymbol{\beta}_j\right) \!= a_i C^y_1 + b_i C^y_2,
    \end{split}
\end{equation}
where the $\mathsf{ind}$ function is for finding the index (i.e., $V_{\mathsf{ind}(V_j)} = V_j$), and $C^y_1$, $C^y_2$ are constants. 
Here, the slope $a_i$ can be expressed as
\begin{equation} \label{eq:a_i}
    a_i = \frac{\tilde{Q}^{\text{B}}_i(V_2) - \tilde{Q}^{\text{B}}_i(V_1)}{V_2-V_1} = \frac{Q^{\text{B}}_i(V_2) - Q^{\text{B}}_i(V_1) + \bar{Q}^{\text{B}}(V_1) - \bar{Q}^{\text{B}}(V_2)}{(V_2-V_1) \max \text{std}(Q^{\text{B}})} = (Q^{\text{B}}_i(V_2) - Q^{\text{B}}_i(V_1)) C^a_1 + C^a_2,
\end{equation}
where $C^a_1$ and $C^a_2$ are constants.
While various metrics can be used for representing the y-intercept $b_i$, we use $\text{mean}(Q^{\text{B}}_i(V_1\text{--}V_2))$ given that $\boldsymbol{\beta}$ is generally flat in each section.
This can be expressed as
\begin{equation} \label{eq:b_i}
    b_i = \text{mean}(Q^{\text{B}}_i(V_1\text{--}V_2))C_1^b + C_2^b,
\end{equation}
where $C^b_1$ and $C^b_2$ are constants.
By combining Equations \ref{eq:linear_approximation}--\ref{eq:b_i}, we conclude that only two features are needed to describe each section: $Q^{\text{B}}_i(V_2)-Q^{\text{B}}_i(V_1)$ and $\text{mean}(Q^{\text{B}}_i(V_1\text{--}V_2))$.

\begin{figure}[h!]
\includegraphics[width=\textwidth]{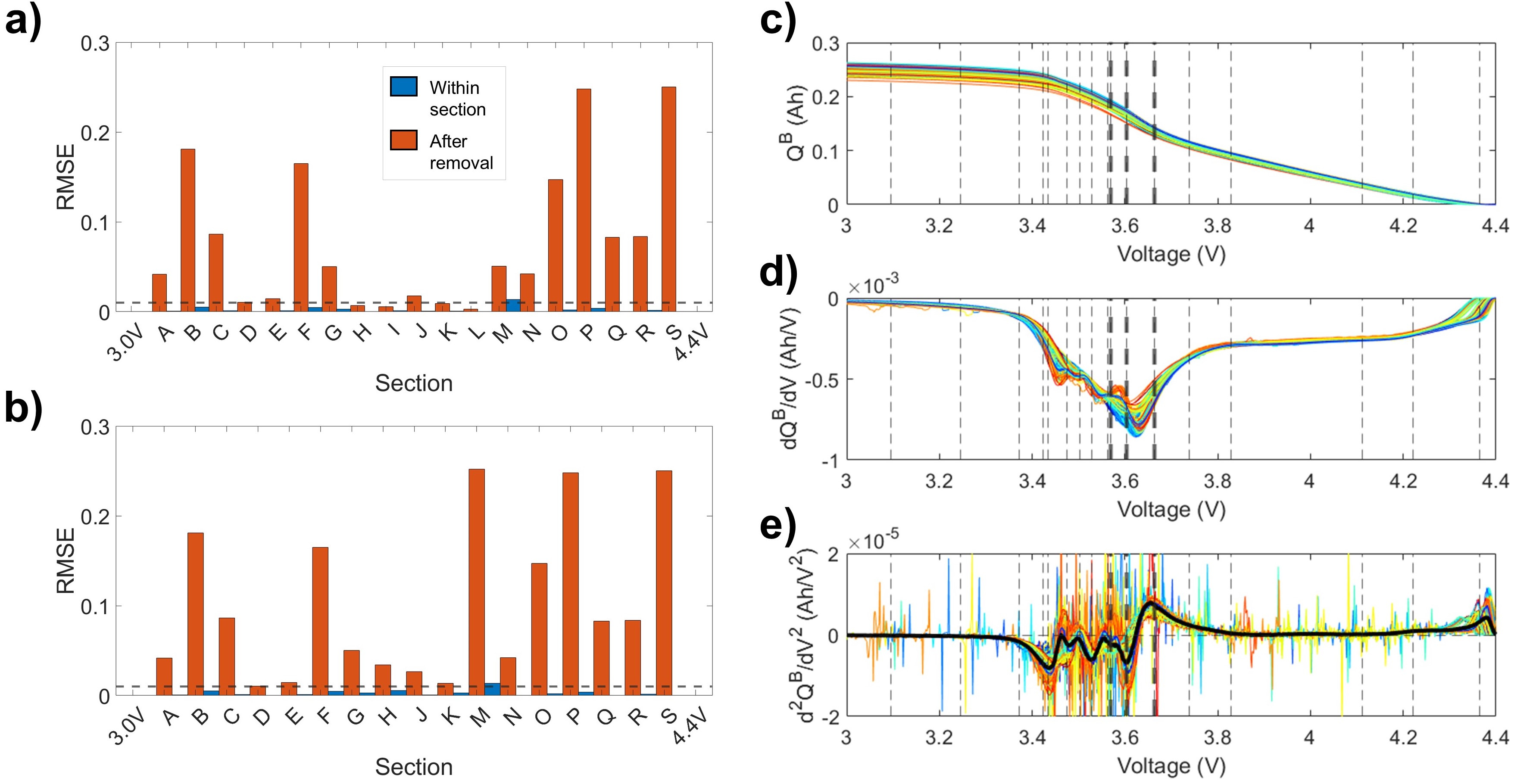}
\centering
\vspace{-0.7cm}
\caption{(a) \gls{rmse} when predicting $\hat{y}_i^{\text{section}}(V_1, V_2)$ with $Q^{\text{B}}_i(V_2)-Q^{\text{B}}_i(V_1)$ and $\text{mean}(Q^{\text{B}}_i(V_1\text{--}V_2))$ in each section in Figure \ref{fig:partitioning}. Blue bars indicate the \gls{rmse} in each section whereas the orange bars indicate the \gls{rmse} when removing such boundary. The horizontal dotted line indicates the threshold for determining which boundary to remove, which was set to 0.01 in this study. (b) \gls{rmse} graph after completing the section merging step. All orange bars exceed the threshold. (c) $Q^{\text{B}}(V)$, (d) $\text{d}Q^{\text{B}}/\text{d}V(V)$, and (e) $\text{d}^2Q^{\text{B}}/\text{d}V^2(V)$ with the vertical dotted lines indicating boundaries selected after Algorithm~\ref{alg:merge_sections} (thin) and the three boundaries near \SI{3.6}{V} (thick). The colors indicate the normalized cycle life and the thick black solid line in panel (e) is for the column-wise average.}
\label{fig:merge_partitions}
\end{figure}

While there are 19 boundaries in Figure \ref{fig:partitioning}, not all of the boundaries might be necessary for partitioning the input data.
For example, boundaries L and M seem to be very close so that using only one might be sufficient.
To determine whether we can remove the boundary $\mathsf{ind}(V_j)$, we can check the average of \gls{rmse} among five inner loops when predicting $\hat{y}_i^{\text{section}}(V_{j-1}, V_{j+1})$ with $Q^{\text{B}}_i(V_{j+1})-Q^{\text{B}}_i(V_{j-1})$ and $\text{mean}(Q^{\text{B}}_i(V_{j-1}\text{--}V_{j+1}))$.
For example, Figure \ref{fig:merge_partitions}a shows that the average \gls{rmse} when removing the boundary L is less than 0.01, which implies that removing the boundary L gives an error less than $\exp(0.01) \approx 1\%$.
Similarly, the boundary I that has the second-least \gls{rmse} can also be removed, which is expected given that boundaries H and I are close to each other in Figure \ref{fig:partitioning}.
The process of removing the boundaries can be iterated until any additional boundary causes more than $1\%$ error in the corresponding section (see Figure \ref{fig:merge_partitions}b) where the algorithm can be found in Algorithm~\ref{alg:merge_sections}.
From Figure \ref{fig:merge_partitions}b, we observe that neither boundaries J nor K can be removed although they seem to be very close to each other in Figure \ref{fig:partitioning}.
This highlights the fact that Algorithm~\ref{alg:merge_sections} is actually sensitive to the information encoded in specific voltage ranges, which will be further discussed later.

\begin{algorithm}[h!]
\caption{Merging partitioned sections}\label{alg:merge_sections}
\begin{algorithmic}
\State $x_0 \gets \{j \in \{1,\cdots{},p-1\}|\text{ jump occurred at $j$}\}$
\State $\text{RMSE}^{\text{merge}} \gets \text{Average of RMSE among five inner loops when removing one boundary}$
\State $\text{th}^{\text{merge}} \gets \text{Threshold for merging the sections}$ \Comment{0.01 in this study.}
\While{$\min(\text{RMSE}^{\text{merge}}) \leq \text{th}^{\text{merge}}$} 
    \State Remove the boundary that corresponds to the minimum $\text{RMSE}^{\text{merge}}$ from $x_0$
    \State Recalculate $\text{RMSE}^{\text{merge}}$ using the updated $x_0$
\EndWhile
\end{algorithmic}
\end{algorithm}

To assign physical meaning to the partitioned sections from Algorithm~\ref{alg:merge_sections}, we plot the final partitioning on top of $Q^{\text{B}}(V)$, $\text{d}Q^{\text{B}}/\text{d}V(V)$, and $\text{d}^2Q^{\text{B}}/\text{d}V^2(V)$ graphs in Figure \ref{fig:merge_partitions}c--e, respectively.
From the figure, it is observed that the boundaries well capture the characteristics of $\text{d}^2Q^{\text{B}}/\text{d}V^2(V)$ curve, especially near the voltage range of $\SI{3.6}{V}$ where the largest peak and valley are identified (see thick vertical dotted lines in Figure \ref{fig:merge_partitions}c--e).

When Algorithm~\ref{alg:merge_sections} yields $N^{\text{sec}}$ sections, we have $2N^{\text{sec}}$ features (difference and mean features in each section) in total.
Among these, we can further down-select the features based on the correlation plot as shown in Algorithm~\ref{alg:downselect_features}.
First, we select the feature with the highest Pearson correlation with the cycle life.
Then, the features that have a high Pearson correlation (e.g., $>0.2$) with the selected features are filtered out to avoid multicollinearity \cite{mei_modeling_2022, hassanien_machine_2019, liu_daily_2020}.
This process can be iterated until there is no feature left with a high Pearson correlation (e.g., $> 0.4$) with the output.
As a result, the two features, $Q^{\text{B}}(\SI{3.57}{V})-Q^{\text{B}}(\SI{3.60}{V})$ and $Q^{\text{B}}(\SI{3.60}{V})-Q^{\text{B}}(\SI{3.66}{V})$, were selected using Algorithm~\ref{alg:downselect_features} for Outer loop 1.
The selected voltage values match with the index for the peak and valley of $\text{d}^2Q^{\text{B}}/\text{d}V^2(V)$ curve in Figure \ref{fig:merge_partitions}e, implying that the designed features capture some physical meaning.
The feature design results on other outer loops are displayed in Table \ref{tab:feature_design_results} (see \gls{si} \ref{appdx:partitioning other_outer} for partitioning results in other outer loops).

\begin{algorithm}[h!]
\caption{Downselection of features based on the correlation matrix}\label{alg:downselect_features}
\begin{algorithmic}
\State $\mathbf{A}^{\text{feat}} \gets \text{A matrix}\in \mathbb{R}^{n\times(2N^{\text{sec}}+1)} \text{ with }2N^{\text{sec}} \text{ features and }\mathbf{y}$
\State $\mathbf{R} \gets \text{Correlation matrix of } \mathbf{A}^{\text{feat}}$
\State $\mathbf{R}^{\text{red}} \gets \mathbf{R}$ \Comment{Will be reduced during the While-loop.}
\State $\text{th}^{\text{PC}, \mathbf{X}} \gets \text{Threshold for Pearson correlation with other features}$ \Comment{0.2 in this study.}
$\text{th}^{\text{PC}, \mathbf{y}} \gets \text{Threshold for Pearson correlation with $\mathbf{y}$}$ \Comment{0.4 in this study.}
\State $\text{ind}^{\mathbf{y}} \gets \text{Indices of } i \in \{1, \cdots{}, \text{nrows}(\mathbf{R}^{\text{red}})-1\} \text{ where } \mathbf{R}^{\text{red}}_{i, \text{nrows}(\mathbf{R}^{\text{red}})} > \text{th}^{\text{PC}, \mathbf{y}}$
\While{$\text{ind}^{\mathbf{y}} \neq \emptyset$} 
    \State $\text{ind}^{\text{selected}} \gets \argmax_{i \in \{1, \cdots{}, \text{nrows}(\mathbf{R}^{\text{red}})-1\}} \mathbf{R}^{\text{red}}_{i, \text{nrows}(\mathbf{R}^{\text{red}})}$
    \State Append $\text{ind}^{\text{selected}}$ to $\text{ind}^{\text{feat}}$
    \State Remove rows/columns with $\mathbf{R}^{\text{red}}_{i, \text{ind}^{\text{selected}}} > \text{th}^{\text{PC}, \mathbf{X}}$ from $\mathbf{R}^{\text{red}}$
    \State $\text{ind}^{\mathbf{y}} \gets \text{Indices of } i \in \{1, \cdots{}, \text{nrows}(\mathbf{R}^{\text{red}})-1\} \text{ where } \mathbf{R}^{\text{red}}_{i, \text{nrows}(\mathbf{R}^{\text{red}})} > \text{th}^{\text{PC}, \mathbf{y}}$
\EndWhile
\end{algorithmic}
\end{algorithm}

\begin{table}[h!]
\begin{center}
\begin{tabular}{|c|c|c|c|}
\hline
Outer & $\lambda$ & Boundaries for final partitioning ($\SI{}{V}$) & Selected features\\ 
 \hline
1 & 0.3603 & \begin{tabular}{@{}c@{}} 3.00    3.10    3.25    3.37    3.42  3.43  3.48  3.50    3.53 3.56 \\  \textbf{3.57}    \textbf{3.60}    \textbf{3.66}   3.74    3.83    4.11    4.22 4.36    4.40\end{tabular} & \begin{tabular}{@{}c@{}}$Q^{\text{B}}(\SI{3.57}{V})-Q^{\text{B}}(\SI{3.60}{V})$ \\ $Q^{\text{B}}(\SI{3.60}{V})-Q^{\text{B}}(\SI{3.66}{V})$\end{tabular}\\ \hline
2 & 0.3925 & \begin{tabular}{@{}c@{}c@{}} 3.00    3.11    3.27    3.37    3.42    3.48    3.50    3.54    \textbf{3.58}\\    \textbf{3.61} \textbf{3.64}    3.73    3.82    3.91    4.22    4.36    4.40\end{tabular} & \begin{tabular}{@{}c@{}}$Q^{\text{B}}(\SI{3.58}{V})-Q^{\text{B}}(\SI{3.61}{V})$ \\ $Q^{\text{B}}(\SI{3.61}{V})-Q^{\text{B}}(\SI{3.64}{V})$\end{tabular}\\ \hline
3 & 0.6788 & \begin{tabular}{@{}c@{}c@{}} 3.00    3.10    3.29    3.37    3.42    3.47    3.50  \textbf{3.57}    \textbf{3.61}\\    \textbf{3.64} 3.73    3.82    3.91    4.21    4.36    4.40\end{tabular} & \begin{tabular}{@{}c@{}}$Q^{\text{B}}(\SI{3.57}{V})-Q^{\text{B}}(\SI{3.61}{V})$ \\ $Q^{\text{B}}(\SI{3.61}{V})-Q^{\text{B}}(\SI{3.64}{V})$\end{tabular}\\ \hline
4 & 0.3631 & \begin{tabular}{@{}c@{}} 3.00    3.10    3.30    3.37  3.46    3.48    3.50    3.54    \textbf{3.58}\\    \textbf{3.60}    \textbf{3.67}    3.73 3.83  3.91  4.01    4.22
4.36    4.40\end{tabular} & \begin{tabular}{@{}c@{}}$Q^{\text{B}}(\SI{3.58}{V})-Q^{\text{B}}(\SI{3.60}{V})$ \\ $Q^{\text{B}}(\SI{3.60}{V})-Q^{\text{B}}(\SI{3.67}{V})$\end{tabular}\\ \hline
5 & 0.5510 & \begin{tabular}{@{}c@{}c@{}} 3.00    3.10    3.26    3.37    3.42    3.48    3.50    \textbf{3.57}    \textbf{3.60}\\    \textbf{3.64} 3.75    3.83    3.92    4.31  4.36  4.40 \end{tabular} & \begin{tabular}{@{}c@{}}$Q^{\text{B}}(\SI{3.57}{V})-Q^{\text{B}}(\SI{3.60}{V})$ \\ $Q^{\text{B}}(\SI{3.60}{V})-Q^{\text{B}}(\SI{3.64}{V})$\end{tabular}\\ \hline
\end{tabular}
\caption{Feature design results for each outer loop when using $Q^{\text{B}}(V)$. The voltage values used in the selected features are marked in bold.}
\label{tab:feature_design_results}
\end{center}
\end{table}

\section{Results and discussion}\label{sec:results_discussion}
\subsection{Evaluation of designed features}\label{sec:results_feature_evaluation}

In this section, we evaluate the designed features by comparing the performance of the agnostic model, autoML model, and the model trained by the designed features (i.e., \textit{designed} model).
Table \ref{tab:model_results} displays the number of features and the performance metrics of the best models when using each approach.
Nonlinear algorithms include \gls{rf}, \gls{svr}, \gls{xgb}, \gls{alven}, and \gls{lcen}, whereas the linear algorithms include \gls{rr}, \gls{en}, \gls{pls}, and \gls{pls}.
\gls{alven} and \gls{lcen} algorithms up to the degree of three \cite{sun_alven_2020} are considered for the agnostic and designed models whereas we only use a degree of one for the autoML models as the $\textsf{tsfresh}$ package already contains various nonlinear transformations.

\begin{table}[h!]
\begin{center}
\footnotesize
\begin{tabular}{|c|c|c|c|c|c|}
\hline
Models & \# features & Median \gls{rmse} & Max \gls{rmse} & Median \gls{mape} & Max \gls{mape}\\ \hline
Agnostic (Nonlinear) & 4 & 107.10 & 124.81 & 11.06 & 11.35\\ 
AutoML (Nonlinear) & 155 & \textbf{86.17} & 136.92 & 9.72 & \textbf{10.85}\\ 
Designed (Nonlinear) & 2 & 97.33 & \textbf{108.54} & \textbf{9.20} & 11.93\\
Agnostic (Linear) & 4 & 134.13 & 189.01 & 14.84 & 15.86\\ 
AutoML (Linear) & 190 & 99.87 & 127.70 & 9.87 & 11.62\\ 
Designed (Linear) & 2 & 113.16 & 123.73 & 11.30 & 14.29\\ \hline
\end{tabular}
\caption{The predictive performance of the best models among 52 agnostic models, 2,448 autoML models, and models using the designed features listed in Table \ref{tab:feature_design_results}, respectively. Bold numbers indicate the smallest value at each column, implying that the model performed best among all approaches.}
\label{tab:model_results}
\end{center}
\end{table}

Table \ref{tab:model_results} shows that the designed model that uses only two features has comparable performance as the autoML model (e.g., each outperforms the other based on two metrics).
This result is especially remarkable given that the designed features overcome the main limitations of other approaches; the agnostic model is trained from the features that directly encode the formation protocol and cannot be used for diagnostic purposes (see Section \ref{sec:methods_data_description}), whereas the autoML model is the extreme case where the interpretability of the model is sacrificed to obtain a better predictive model.
On the other hand, the designed model uses two simple $Q(V)$ features that can be applied to formation protocols that do not necessarily follow the template used by Cui et al.~\cite{cui_data-driven_2024}.
These facts provide the designed model with both high flexibility and interpretability, which makes it appropriate for optimizing the formation protocol.

We attribute the outstanding performance of the designed features over autoML features to the fact that our partitioning algorithm well identifies specific voltage ranges with critical information (see Figures \ref{fig:partitioning} and \ref{fig:merge_partitions}c--e).
The features generated from the autoML approach typically use the entire dataset or subset of the dataset that is chosen based on simple statistics (e.g., percentile or arbitrarily chosen thresholds).
On the other hand, the features designed from our framework use the subset that is determined by considering the relationship between the input data and the output (e.g., jumps occurred within $\boldsymbol{\beta}$ or Algorithm~\ref{alg:merge_sections}).
Since the designed features use a subset of input data with similar information once at a time, they are sensitive to voltage-specific information, ultimately leading to a high predictive power.

\begin{table}[h!]
\begin{center}
\footnotesize
\begin{tabular}{|c|c|c|c|c|}
\hline
Models & $\Delta$Median \gls{rmse} & $\Delta$Max \gls{rmse} & $\Delta$Median \gls{mape} & $\Delta$Max \gls{mape}\\ \hline
Agnostic (Linear - Nonlinear) & 27.03 & 64.21 & 3.77 & 4.51\\ 
Designed (Linear - Nonlinear) & 15.83 & 15.18 & 2.11 & 2.36\\ \hline
\end{tabular}
\caption{The difference between the nonlinear and linear models in four performance metrics for the agnostic and designed approaches as shown in Table \ref{tab:model_results}.}
\label{tab:model_results_difference}
\end{center}
\end{table}

From Table \ref{tab:model_results_difference}, we observe that the performance gap between the nonlinear- and linear-designed models is around two-thirds and less than half of the agnostic model in median and maximum metrics, respectively.
While a large gap between the nonlinear and linear models for the agnostic approach was expected from the data interrogation results (see Figure \ref{fig:data_description}c--e), the designed model having a smaller gap between the nonlinear and linear models implies that the proposed framework well captures the linear correlation between the output and the designed features.
This result is consistent with the proposed framework using linear methods for important steps such as finding $\boldsymbol{\beta}$, partitioning the input data, merging the sections, and down-selecting the features.
Given that the designed features show low variation ($\sim$12\%) across 62 different protocols and simplify the regression model, they are suitable for performing extreme early cycle life prediction for untrained formation protocols (i.e., extrapolation) and thus for optimizing the formation process.

\subsection{Physical meaning of the designed features}\label{sec:discussion_autoML_model}

In this section, we investigate the physical origins of the designed features to understand their outstanding performance over features from agnostic and autoML approaches.
We aim to (1) develop a foundational understanding of the significance of these features for a subset of cells and (2) explain why the designed features demonstrate more robust performance compared to other data-driven modeling approaches.
As shown in Figure \ref{fig:merge_partitions}c--e, the designed features do not directly correspond to features in the dataset's average discharge capacity or differential capacitance curves, which have been widely used in previous studies as indicators of lifetime \cite{severson_data-driven_2019, pinson_theory_2013, attia_electrochemical_2019, das_electrochemical_2019}.
However, the voltage windows identified from our framework show a strong overlap with local maxima and minima in the dataset's average second derivative of capacity (d$^2Q$/d$V^2$) data, particularly in the range of 3.4 to $\SI{3.7}{V}$ where the downselected features in Table \ref{tab:feature_design_results} are concentrated.
This alignment between key curve-defining features of the dataset like the maximum and minima curvature regions and the selected voltage points highlights the high interpretability of the designed features.

Given the wide variations in the formation protocols within the dataset, precisely interpreting the physical meaning of these features is challenging.
Therefore, we explore the features specific to slow formation for two reasons.
First, a majority of the cells in the studied dataset are under relatively slow formation conditions, so we expect that the designed features may have a preference to the physical phenomena driving performance in slow formed cells.
Second, the physical understanding during `high performing' slow formation is more established relative to that for fast formation.
For slowly formed Li-battery systems, it has long been hypothesized that \gls{sei} growth is the dominant mechanism for capacity fade,
which becomes self-limiting at long times due to diffusion limitation of solvent molecules across the \gls{sei} layer \cite{peled_role_1995,ploehn_solvent_2004,ramadass_development_2004,smith_high_2011,peled_reviewsei_2017,wood_formation_2019,weng_modeling_2023,an_state_2016}. 
A dynamical transition from reaction-limited to diffusion-limited \gls{sei} growth has also been proposed~\cite{pinson_theory_2013}, roughly corresponding to classical two-layer models of \gls{sei} growth, in which the rapid electrodeposition of a dense, inorganic primary \gls{sei} layer is followed by slow, diffusion-limited growth of a thick, porous secondary \gls{sei} layer~\cite{thevenin_impedance_1987,aurbach_correlation_1994,peled_composition_2001,peled_reviewsei_2017}.  
Recently, there has been growing evidence of two growth regimes during the formation process with significant portion of \gls{sei}-related capacity generated during the early stage \cite{huang_evolution_2019,attia_electrochemical_2019,das_electrochemical_2019,weng_modeling_2023,thaman_two-stage_2024,von_kolzenberg_solidelectrolyte_2020}.
For slow formation, the primary \gls{sei} is suspected to be \textit{well-formed} where, by the end of formation, the system is saturated with degradation products, and the total lithium consumed by \gls{sei} production is similar across formation protocols. 

\begin{figure}[h!]
\includegraphics{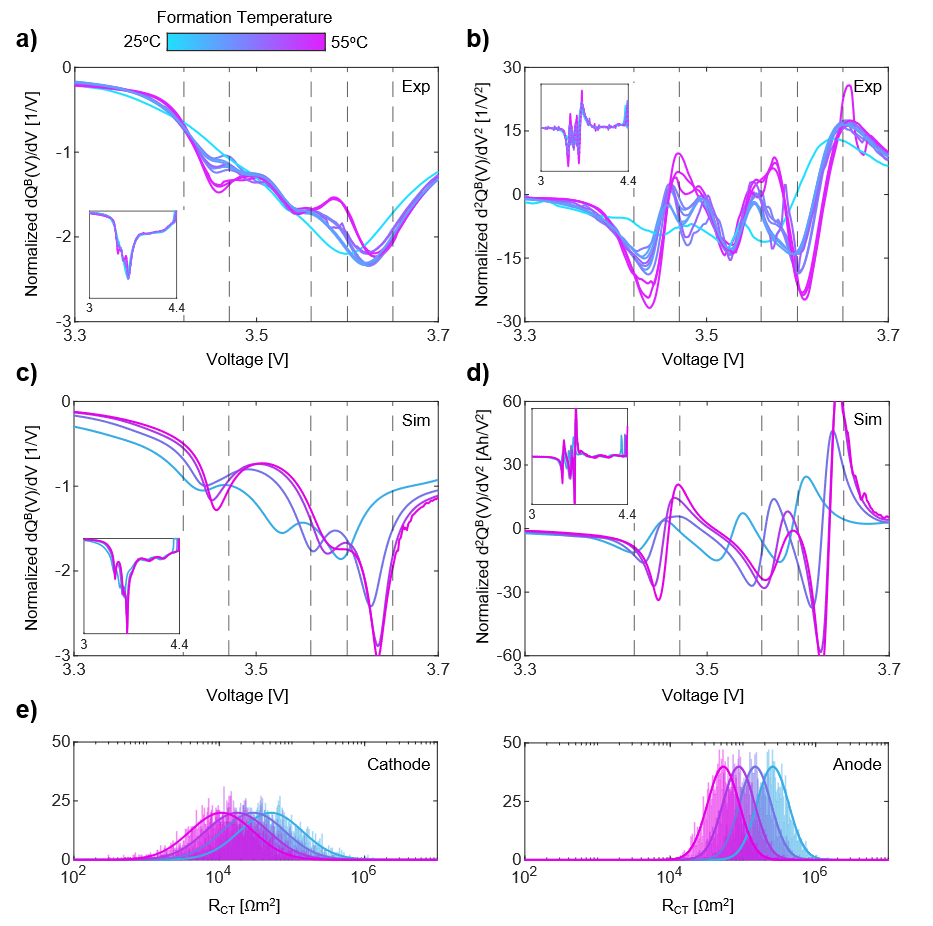}
\centering
\vspace{-0.3cm}
\caption{Measured (a) differential capacitance and (b) d$^2Q$/d$V^2$ for `extreme slow formation' cells across formation temperatures. Reactive particle ensemble simulated (c) differential capacitance and (d) d$^2Q$/d$V^2$ across formation temperatures. (e) Simulated distribution of particle charge transfer resistance across formation temperatures. Both electrodes reaction rates are governed by a singular log-Gaussian basis where at 298 K, the model is parameterized by $E_\text{A} = 45$ kJ/mole, $\bar{k}_{0,\text{c}} = 5\times10^{-7}$ A/m$^2$s, $\bar{k}_{0,\text{a}} = 10^{-7}$ A/m$^2$s, $\sigma_\text{c} = 1$, and $\sigma_\text{a} = 0.5$. Full model details can be found in Section \ref{si-model} of the SI.}
\label{fig:dQdV_physics_model}
\end{figure}

To investigate this physical description of slow formation, we define the \gls{esf} subset, consisting of 32 out of 178 cells, by the condition $CC_1 < 0.05 \text{C}$.
In post-formation C/20 low-rate test (\gls{rpt}) across these cells, shown in the SI Section \ref{sec:fit_utilization}, the differential capacitance and d$^2Q$/d$V^2$ electrochemical signatures are nearly indistinguishable.
This is an indication that the electrode utilization and remaining lithium inventory are likely similar across these cells.
However, the performance, as quantified through cycle life, is not identical across all these cells.
Specifically, the cells formed at the higher formation temperature tested of 55$^\circ$C perform significantly better than the rest, perhaps indicating that the \textit{quality} of formed degradation layers like \gls{sei} may be temperature dependent.
The impact of the formation temperature on the electrochemical signature can be seen directly in the formation operating datasets, such as the d$Q$/d$V$ and d$^2Q$/d$V^2$ curves in Step B as shown in Figure \ref{fig:dQdV_physics_model}ab.
These curves show little variation within cells formed at the same temperature but large variations between different formation temperatures.
As the formation temperature decreases, a smoothing of both curves is observed, which makes certain agnostic model feature visible through the electrochemical signatures. 

These trends are explained through the use of a surrogate model for the full cell system, where the focus is on the temperature dependence of the electrochemical signatures during a simulation of Step B.
The underlying coupled transport and reaction dynamics for each electrode is simplified to a large ensemble of reactive particles, where each particle has a rate constant sampled from a defined distribution.
The model can also be understood in the context of a distributed resistance model, where many transport and reaction resistances are being lumped into the charge transfer resistance, $R_{\text{CT}} = k_\text{B}T/ek_0$.
The calculated current density in the underlying microscopic model for the $i$th particle in electrode $j$ is modeled as
\begin{gather}
    j(\bar{c}^{(i)}_j, V, T) = k_{0,j}^{(i)} \exp\!{\left(\frac{-E_\text{A}}{R} \!\left(\frac{1}{T} - \frac{1}{298} \right)\!\right)} (1-\bar{c}^{(i)}_j) \sqrt{\bar{c}^{(i)}_j} \sinh\!{ \left(- \frac{e \eta(\bar{c}^{(i)}_j, V_j)}{k_\text{B} T}\right)},
\end{gather}
based on the quantum theory of \gls{icet}~\cite{bazant_unified_2023}. The \gls{icet} rate expression has a symmetric Butler-Volmer dependence on overpotential (for equal reduction and oxidation ion-transfer energies) and an asymmetric dependence on electrode filling fraction, recently confirmed by learning from X-ray images of battery nanoparticles~\cite{zhao_learning_2023}. 
The model also includes an Arrhenius temperature dependence in the pre-factor with an activation barrier, $E_{\text{A}}$ (the activation enthalpy of \gls{icet}), which captures the effects of the formation protocol temperature on the simulated dynamics.
The full model formulation is described in Section \ref{si-model} of the SI.

By simulating the system with a log-Gaussian distribution for particle charge transfer resistance at each electrode, characterized by a mean and standard deviation, we are able to simulate a set of Step B curves across different temperatures with five free parameters.
Comparing the simulated results in Figure \ref{fig:dQdV_physics_model}c--d with the experimental dataset in Figure \ref{fig:dQdV_physics_model}a--b, we see striking similarities between the shape of the underlying features where the location of the local maxima and minima in the d$^2Q$/d$V^2$ and their trends with temperature.
These similarities allow us to theorize that the features contain information both on the average resistance of the two electrodes and the underlying resistance heterogeneity from the complex underlying microscopic system state. 
While the former may be gleaned from simpler features such as the differential resistance at a given \gls{soc}, the latter is unique to the designed features and could lead to the high performance observed.
This underlying heterogeneity is likely a consequence of the quality of the \gls{sei} being formed, which is not obviously extractable from just the agnostic model parameters.
In all, we hypothesize that the designed features perform better than the agnostic model because their values not only encode some of the agnostic model parameters, such as formation protocol temperature, but also encode some an underlying heterogeneity in the microscopic particle resistances which varies from cell-to-cell and translates into the electrochemical signatures during formation.

\section{Conclusion}\label{sec:conclusions}

In this work, we develop a systematic feature design framework that enables extreme early cycle life prediction with minimal domain knowledge and user input.
The main strength of our framework over hand-crafted features is that it explores potential information encoded in every available input data type, rather than being limited by existing domain knowledge.
Using our framework, two simple $Q(V)$ features are designed from Step B, which is the last discharge step of the formation protocol. 
These features do not require additional diagnostic cycles and they showed good performance in comparing formation protocols with different temperatures, which is known as the key parameter affecting the quality of \gls{sei} layers.
Our designed features result in $9.20\%$ median \gls{mape}, which outperforms the best model among all 2,448 autoML models constructed in this work that used hundreds of features.
This fact highlights that a careful feature engineering strategy used in our framework can greatly reduce the number and complexity of features while retaining the model performance.
The designed features are consistent among all five outer loops, indicating that our work is robust with respect to different training-test splits.
Our framework provides high interpretability to the designed features since the main processes of the framework are based on linear methods.
As such, we used a multi-particle physics-based model to understand the physical meaning encoded in the designed features.
From physics-based investigation, we attribute the predictiveness of the designed $Q(V)$ features to its ability to learn the formation temperature, the most important feature in the agnostic model. 
Additionally, the designed features are indicative of the underlying features of the $\text{d}Q/\text{d}V$ and $\text{d}^2Q/\text{d}V^2$, which can be attributed to a distribution of particle resistances. 
This underlying heterogeneity encoded in the electrochemical cycling data cannot be captured from the features used in the agnostic model.
We believe that the proposed feature design framework can be used to accelerate lithium-ion battery research by leveraging the interplay between the systematic data-driven approach and mechanistic understanding for formation optimization.

\section*{CRediT authorship contribution statement} \label{sec:CRediT}
\textbf{Jinwook Rhyu: }Conceptualization, Methodology, Formal Analysis, Investigation, Data curation, Software, Visualization, Validation, Writing--original draft, Writing--review \& editing. \textbf{Joachim Schaeffer: }Conceptualization, Methodology, Formal Analysis, Investigation, Software, Writing--review \& editing. \textbf{Michael L. Li: }Formal Analysis, Investigation, Software, Visualization, Writing--original draft, Writing--review \& editing. \textbf{Xiao Cui: }Formal Analysis, Investigation, Data curation, Writing--review \& editing. \textbf{William C. Chueh: }Supervision, Writing--review \& editing, Funding acquisition. \textbf{Martin Z. Bazant: }Supervision, Writing--review \& editing, Funding acquisition.  \textbf{Richard D. Braatz: }Conceptualization, Methodology, Investigation, Supervision, Writing--review \& editing, Funding acquisition.

\section*{Data and Code Availability} \label{sec:data_code_availability}
The raw data used in this work can be found at https://data.matr.io/8/~\cite{cui_data-driven_2024}.
The code used in this work can be found in https://github.com/JinwookRhyu/Systematic\_Feature\_Design\_Framework\_Formation.

\section*{Acknowledgements}
This work was supported by the Toyota Research Institute through D3BATT: Center for Data-Driven Design of Li-Ion Batteries. The authors are grateful to Vivek Lam, Alexis Geslin, Justin Rose, Aki Takahashi, Huada Lian, Elia Arnese-Feffin, and Zhouhang (Amelia) Dai for providing comments.

\newpage
\printbibliography

\newpage
\appendix

\setcounter{figure}{0}
\setcounter{table}{0}
\setcounter{equation}{0}
\setcounter{section}{0}
\renewcommand{\thesection}{S\arabic{section}}
\renewcommand{\thefigure}{S\arabic{figure}}
\renewcommand{\thetable}{S\arabic{table}}
\renewcommand{\theequation}{S\arabic{equation}}
\section{Performance of \texorpdfstring{$R_{\text{LS}}$}{TEXT} in the dataset in Ref.~\cite{cui_data-driven_2024}}\label{appdx:R_LS}

\begin{figure}[h!]
\includegraphics[width=\textwidth]{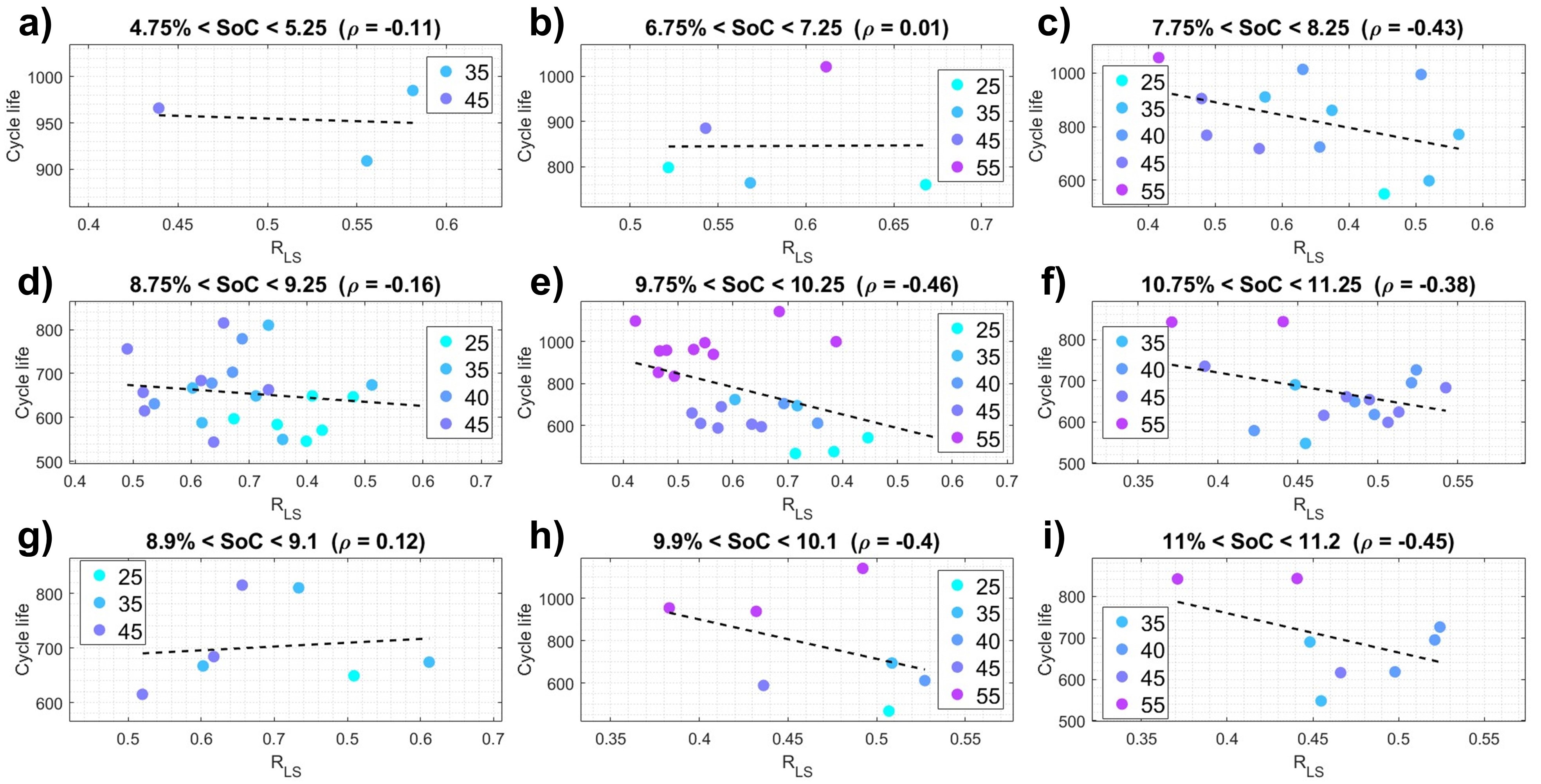}
\centering
\vspace{-0.3cm}
\caption{Linear correlation between $R_{\text{LS}}$ and cycle life in dataset in Ref.~\cite{cui_data-driven_2024}. Each panel includes the data points where the $R_{\text{LS}}$ was measured at \gls{soc} within 0.25$\%$ \gls{soc} error near (a) 5$\%$ \gls{soc}, (b) 7$\%$ \gls{soc}, (c) 8$\%$ \gls{soc}, (d) 9$\%$ \gls{soc}, (e) 10$\%$ \gls{soc}, and (f) 11$\%$ \gls{soc}, and within 0.10$\%$ \gls{soc} error near (g) 9$\%$ \gls{soc}, (h) 10$\%$ \gls{soc}, and (i) 11.1$\%$ \gls{soc}.}
\label{fig:R_LS}
\end{figure}

Figure \ref{fig:R_LS} shows the scatter plot of $R_{\text{LS}}$ and cycle life at specific \gls{soc} levels. 
While the \gls{hppc} protocol used in Ref.~\cite{weng_predicting_2021} applied pulses per every 4$\%$ \gls{soc}, the \gls{hppc} protocol used in Ref.~\cite{cui_data-driven_2024} applied pulses per every 20$\%$ \gls{soc}.
Due to the large \gls{soc} gap between the pulses and asymmetric behavior of $R_{\text{LS}}$ as a function of \gls{soc} \cite{weng_predicting_2021}, using interpolation to estimate the resistance at other \gls{soc} values would be inaccurate for this dataset.
Thus, the data points were grouped by the \gls{soc} values where the $R_{\text{LS}}$ was measured in Figure \ref{fig:R_LS} to avoid errors generated from the interpolation. 
\gls{soc} was calculated based on the capacity measurement during \gls{hppc} cycle:
\begin{equation} \label{eq:SoC}
    \text{SoC} = \frac{Q_{\text{ch}}\text{ until the last pulse} - Q_{\text{dis}}\text{ until the last pulse}}{Q_{\text{ch}}\text{ during charge CCCV}} \times 100\%,
\end{equation}
where $Q_{\text{ch}}$ and $Q_{\text{dis}}$ stand for charge and discharge capacity, respectively.

It can be observed from Figure \ref{fig:R_LS}a--f that the negative correlation between $R_{\text{LS}}$ and the cycle life does not appear for the dataset in Ref.~\cite{cui_data-driven_2024} in \gls{soc} values spanning from 5$\%$ to 11$\%$.
The same observation holds even when narrowing down the \gls{soc} range from 0.5$\%$ to 0.2$\%$ in Figure \ref{fig:R_LS}g--i.
This can be explained by (1) $R_{\text{LS}}$ cannot be used for comparing formation protocols with different formation temperatures or (2) much more careful \gls{soc} measurement is required, where either limits using $R_{\text{LS}}$ for optimizing formation.

\section{Data interrogation results}\label{appdx:data_interrogation}

\begin{figure}[h!]
\includegraphics[width=\textwidth]{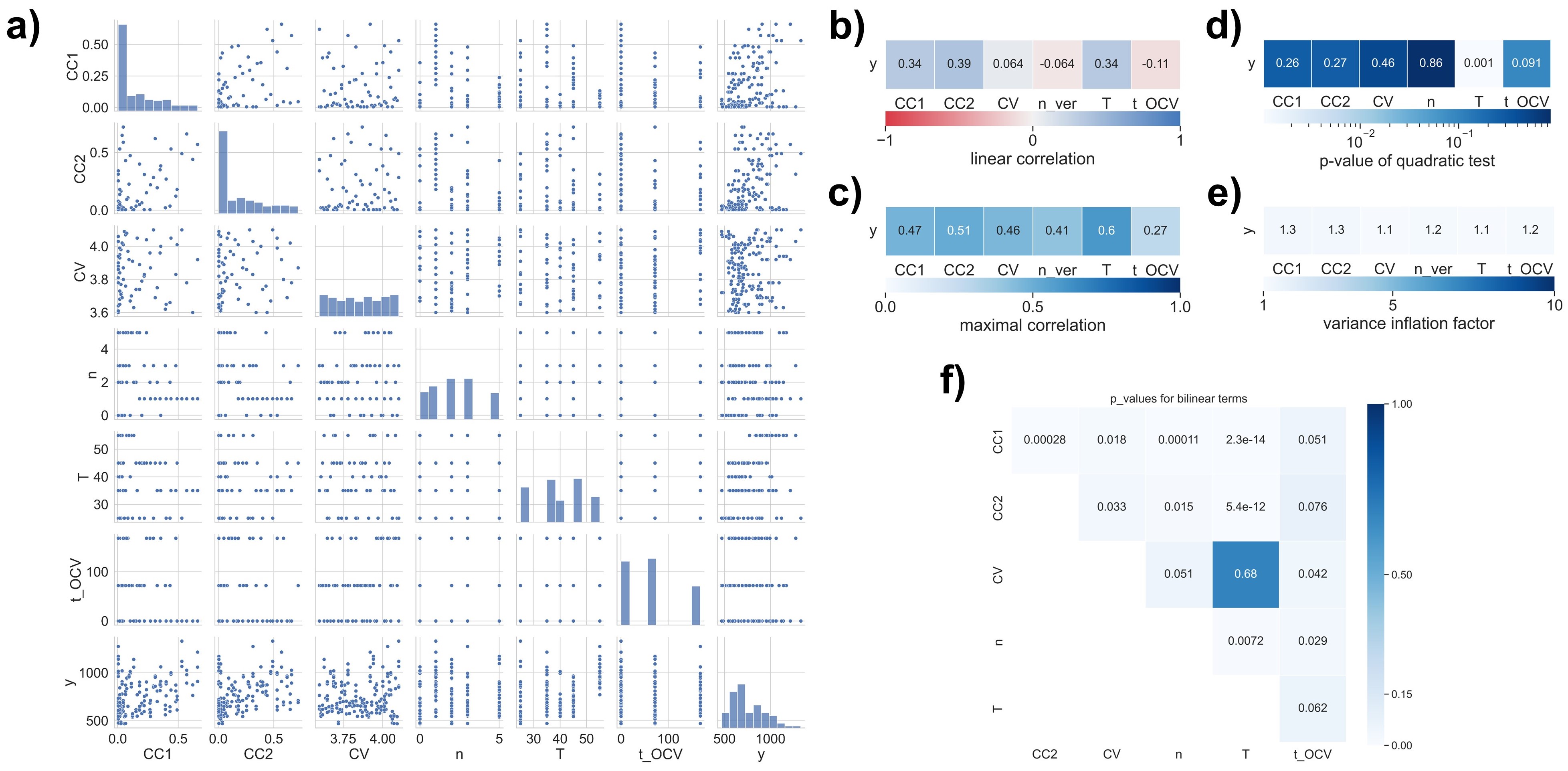}
\centering
\caption{
Data interrogation results using the six formation protocol parameters and cycle life.
(a) Pair plot of six formation protocol parameters and cycle life. (b) Linear correlation (from Pearson correlation) and (c) Maximal correlation (from alternating conditional expectations) between formation protocol parameters and cycle life. (d) Quadratic test with the null hypothesis $H_0: y = w_1x + w_0$ and the alternative hypothesis $H_\text{a}: y = w_2x^2 + w_1x + w_0$. (e) Multicollinearity test with variance inflation factor. (f) Bilinear test with the null hypothesis $H_0: y = w_1x_1 + w_2x_2 + w_0$ and the alternative hypothesis $H_\text{a}: y = w_1x_1 + w_2x_2 + w_{12}x_1x_2 + w_0$.
}
\label{fig:data_description}
\end{figure}

Data interrogation provides preliminary information on the characteristics of the dataset.
Figure \ref{fig:data_description} displays the data interrogation results using the six formation protocol parameters and cycle life.
Comparison of the linear Pearson with maximal correlation coefficients from \gls{ace} algorithm~\cite{breiman_estimating_1985} (Figure \ref{fig:data_description}bc) indicates that there are nonlinear relationships between the six parameters and cycle life.
The maximal correlation coefficients are $\sim$2- to $\sim$7-fold larger than the linear correlation coefficients for the last four parameters, $CV$, $n_{\text{ver}}$, $T$, and $t_{\text{OCV}}$, that is, the nonlinear correlations are greater or equal to linear correlations.
The quadratic test (Figure \ref{fig:data_description}d), which evaluates the null hypothesis of a linear relationship against the alternative of a quadratic relationship, indicates that the cycle life might be related to $T^2$.
The bilinear test result (Figure \ref{fig:data_description}e) shows that the bilinear terms such as $CC_1 \times T$ and $CC_2 \times T$ might be related to the cycle life.
Lastly, having the \gls{vif} values (Figure \ref{fig:data_description}e) close to one implies that the multicollinearity is negligible \cite{obrien_caution_2007}, which is due to the \gls{lhs}.

\section{Detailed split for outer and inner loops}\label{appdx:detailed_split}
The formation protocols were labeled from 1 to 62 in increasing order of $CC_1$, then by increasing order of $CC_2$, followed by increasing order of $CV$, $n_{\text{ver}}$, $T$, and $t_{\text{OCV}}$.
Then, the protocol labels were split into five groups for outer loops after the shuffle using the $\mathsf{numpy}$ package.
For the inner loop, the protocol labels were reassigned to an increasing order of previous labels and then grouped into five based on the remainder of the label divided by five.
For example, the protocol labels in the training set of Outer loop 2 start with $1, 2, 3, 6, 8, 10, \cdots$.
Now, we give them the new labels starting from 1.
Then, Protocol 10 which now has a new label 6 belongs to the Inner Group 1 in Outer loop 1, indicating that Protocol 10 is in the validation set for Inner loop 1 and in the training set for other inner loops.

\begin{table}[h!]
\begin{center}
\tiny
\begin{tabular}{|c|c|c|c|c|c|c|c||c|c|c|c|c|c|c|c|}
\hline
Label & \begin{tabular}{@{}c@{}}$CC_1$ \\ ($\SI{}{A}$)\end{tabular} & \begin{tabular}{@{}c@{}}$CC_2$ \\ ($\SI{}{A}$)\end{tabular} & \begin{tabular}{@{}c@{}}$CV$ \\ ($\SI{}{V}$)\end{tabular} & $n_{\text{ver}}$ & \begin{tabular}{@{}c@{}}$T$ \\ ($^{\circ}$C)\end{tabular} & \begin{tabular}{@{}c@{}}$t_{\text{OCV}}$ \\ ($\SI{}{s}$)\end{tabular} & \begin{tabular}{@{}c@{}}Outer \\ Group\end{tabular} & Label & \begin{tabular}{@{}c@{}}$CC_1$ \\ ($\SI{}{A}$)\end{tabular} & \begin{tabular}{@{}c@{}}$CC_2$ \\ ($\SI{}{A}$)\end{tabular} & \begin{tabular}{@{}c@{}}$CV$ \\ ($\SI{}{V}$)\end{tabular} & $n_{\text{ver}}$ & \begin{tabular}{@{}c@{}}$T$ \\ ($^{\circ}$C)\end{tabular} & \begin{tabular}{@{}c@{}}$t_{\text{OCV}}$ \\ ($\SI{}{s}$)\end{tabular} & \begin{tabular}{@{}c@{}}Outer \\ Group\end{tabular}\\ \hline
1 & 0.0048 & 0.0048 & 3.88 & 0 & 40 & 0 & 1 & 32 & 0.084 & 0.108 & 3.97 & 5 & 25 & 168 & 2\\
2 & 0.0048 & 0.0048 & 3.94 & 5 & 40 & 72 & 4 & 33 & 0.0936 & 0.2256 & 3.92 & 3 & 55 & 72 & 3\\
3 & 0.0048 & 0.0216 & 3.61 & 2 & 40 & 72 & 4 & 34 & 0.1032 & 0.4728 & 3.67 & 0 & 40 & 0 & 2\\
4 & 0.0048 & 0.0216 & 3.64 & 2 & 40 & 72 & 2 & 35 & 0.1128 & 0.0216 & 3.7 & 5 & 55 & 72 & 2\\
5 & 0.0048 & 0.0216 & 3.78 & 2 & 40 & 72 & 2 & 36 & 0.1124 & 0.0048 & 3.91 & 5 & 25 & 72 & 4\\
6 & 0.0048 & 0.0216 & 4.06 & 0 & 40 & 72 & 1 & 37 & 0.132 & 0.0048 & 3.63 & 2 & 55 & 0 & 3\\
7 & 0.0048 & 0.036 & 3.9 & 3 & 40 & 0 & 2 & 38 & 0.1512 & 0.036 & 3.74 & 2 & 45 & 0 & 4\\
8 & 0.0048 & 0.0792 & 3.81 & 3 & 40 & 0 & 5 & 39 & 0.1608 & 0.0792 & 4.05 & 5 & 35 & 72 & 5\\
9 & 0.0048 & 0.2544 & 3.86 & 0 & 40 & 72 & 2 & 40 & 0.18 & 0.0504 & 3.89 & 5 & 45 & 72 & 4\\
10 & 0.0048 & 0.312 & 4.03 & 3 & 40 & 168 & 1 & 41 & 0.18 & 0.4 & 3.74 & 1 & 25 & 0 & 3\\
11 & 0.0168 & 0.3408 & 3.6 & 3 & 40 & 72 & 5 & 42 & 0.1992 & 0.0648 & 3.79 & 0 & 45 & 72 & 3\\
12 & 0.0168 & 0.0048 & 3.65 & 2 & 40 & 168 & 3 & 43 & 0.2184 & 0.0048 & 3.82 & 3 & 45 & 72 & 5\\
13 & 0.0168 & 0.0504 & 3.73 & 5 & 40 & 168 & 1 & 44 & 0.22 & 0.31 & 3.83 & 1 & 25 & 0 & 4\\
14 & 0.0168 & 0.0504 & 3.84 & 2 & 40 & 72 & 2 & 45 & 0.2376 & 0.1512 & 4.04 & 5 & 45 & 168 & 3\\
15 & 0.0168 & 0.0648 & 3.93 & 0 & 40 & 72 & 5 & 46 & 0.2664 & 0.0048 & 3.96 & 2 & 45 & 168 & 2\\
16 & 0.0168 & 0.1368 & 3.95 & 3 & 40 & 0 & 3 & 47 & 0.27 & 0.22 & 3.87 & 1 & 45 & 0 & 4\\
17 & 0.0168 & 0.18 & 3.68 & 2 & 40 & 168 & 1 & 48 & 0.2952 & 0.2832 & 4.1 & 3 & 25 & 168 & 1\\
18 & 0.0168 & 0.5304 & 3.99 & 0 & 40 & 168 & 2 & 49 & 0.31 & 0.62 & 3.69 & 1 & 45 & 0 & 1\\
19 & 0.0264 & 0.0048 & 3.71 & 3 & 40 & 72 & 5 & 50 & 0.3264 & 0.1944 & 3.75 & 2 & 35 & 72 & 4\\
20 & 0.0264 & 0.0048 & 3.98 & 3 & 40 & 72 & 2 & 51 & 0.35 & 0.35 & 3.96 & 1 & 45 & 0 & 1\\
21 & 0.0264 & 0.036 & 3.72 & 2 & 40 & 0 & 5 & 52 & 0.3552 & 0.384 & 4 & 0 & 35 & 168 & 5\\
22 & 0.0264 & 0.1656 & 3.77 & 2 & 40 & 0 & 4 & 53 & 0.3936 & 0.0216 & 3.62 & 3 & 45 & 72 & 2\\
23 & 0.036 & 0.0048 & 3.69 & 0 & 40 & 72 & 5 & 54 & 0.4 & 0.27 & 3.65 & 1 & 25 & 0 & 3\\
24 & 0.036 & 0.648 & 4.02 & 2 & 40 & 0 & 4 & 55 & 0.432 & 0.036 & 3.66 & 2 & 25 & 72 & 2\\
25 & 0.0456 & 0.0216 & 4.01 & 5 & 40 & 0 & 3 & 56 & 0.44 & 0.53 & 4.05 & 1 & 35 & 0 & 4\\
26 & 0.0456 & 0.588 & 3.8 & 3 & 40 & 168 & 1 & 57 & 0.48 & 0.1224 & 3.83 & 3 & 35 & 168 & 1\\
27 & 0.0456 & 0.72 & 3.87 & 3 & 40 & 72 & 3 & 58 & 0.49 & 0.18 & 4.01 & 1 & 45 & 0 & 1\\
28 & 0.0552 & 0.0048 & 4.09 & 0 & 40 & 72 & 3 & 59 & 0.53 & 0.49 & 4.1 & 1 & 25 & 0 & 3\\
29 & 0.0648 & 0.0048 & 4.07 & 3 & 40 & 168 & 5 & 60 & 0.57 & 0.66 & 3.78 & 1 & 35 & 0 & 1\\
30 & 0.0648 & 0.4296 & 3.76 & 5 & 40 & 72 & 4 & 61 & 0.62 & 0.44 & 3.6 & 1 & 35 & 0 & 5\\
31 & 0.0744 & 0.0936 & 4.08 & 2 & 40 & 168 & 1 & 62 & 0.66 & 0.57 & 3.92 & 1 & 35 & 0 & 5\\ \hline

\end{tabular}
\caption{Labels and outer loop split for 62 formation protocols.}
\label{tab:split_formation_protocols}
\end{center}
\end{table}

\newpage
\section{Promisingness of input data based on autoML models}
\label{appdx:promisingness_inputdata_autoML}

\begin{figure}[h!]
\includegraphics[width=\textwidth]{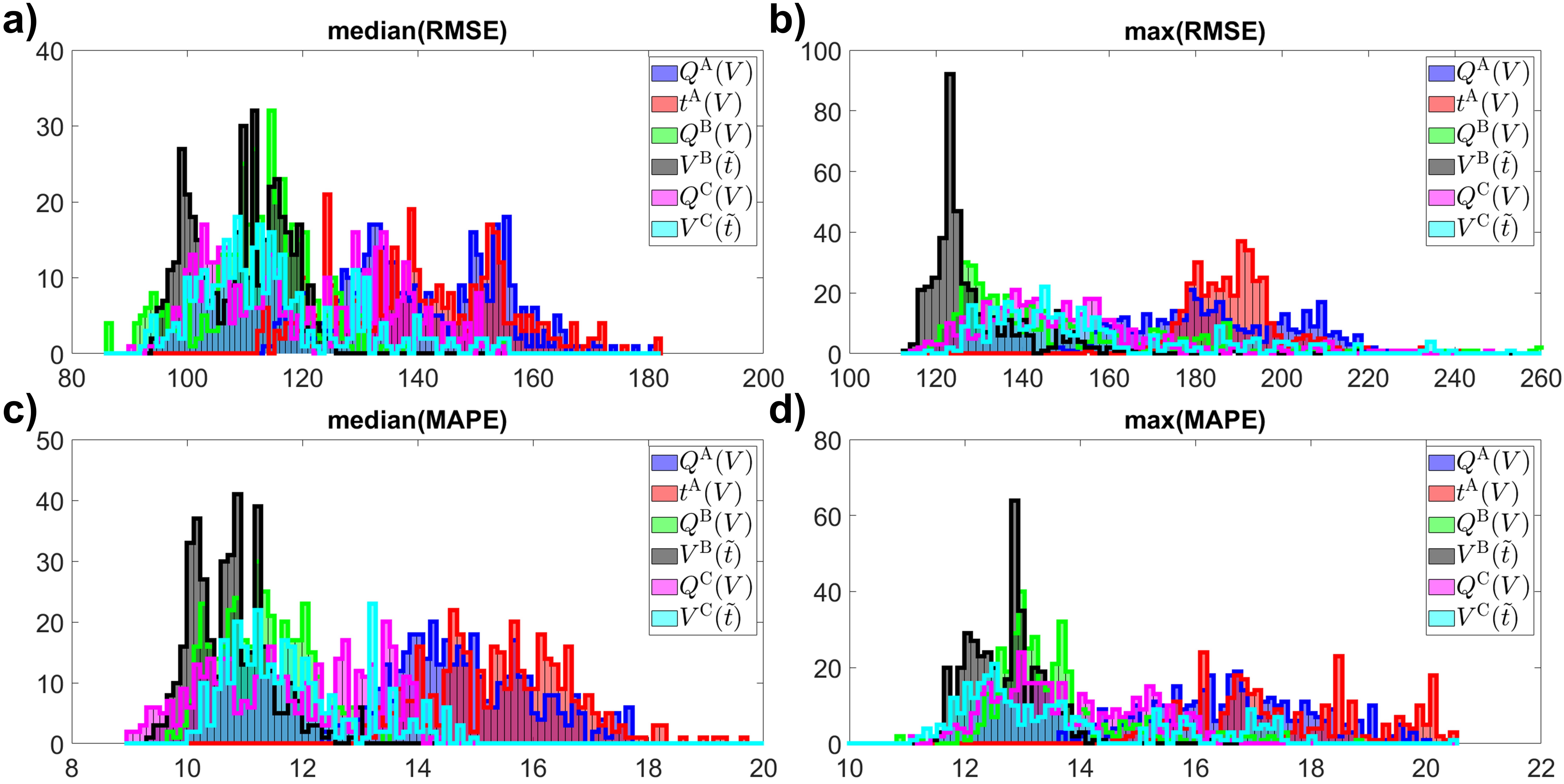}
\centering
\vspace{-0.3cm}
\caption{Histogram of performances of autoML models in (a) median \gls{rmse}, (b) max \gls{rmse}, (c) median \gls{mape}, and (d) max \gls{mape} metrics.}
\label{fig:histogram_autoML}
\end{figure}

Figure \ref{fig:histogram_autoML} displays the histogram of performances of all autoML models constructed in this work in both \gls{rmse} and \gls{mape} metrics.
The input data with distributions at the left (i.e., smaller \gls{rmse} or \gls{mape} metrics) are likely to contain hidden information and thus would possibly lead to promising features.
From Figure \ref{fig:histogram_autoML}, we observe that the four input data types from Steps B and C (green, black, violet, and cyan) are more promising than the two input data types from Step A (blue and red).

\newpage
\section{Partitioning of $Q^{\text{B}}(V)$ in other outer loops}
\label{appdx:partitioning other_outer}

\begin{figure}[htbp!]
\centering
\subfigure[$Q^{\text{B}}(V)$ for Outer loop 2.]{
\includegraphics[width=.470\textwidth]{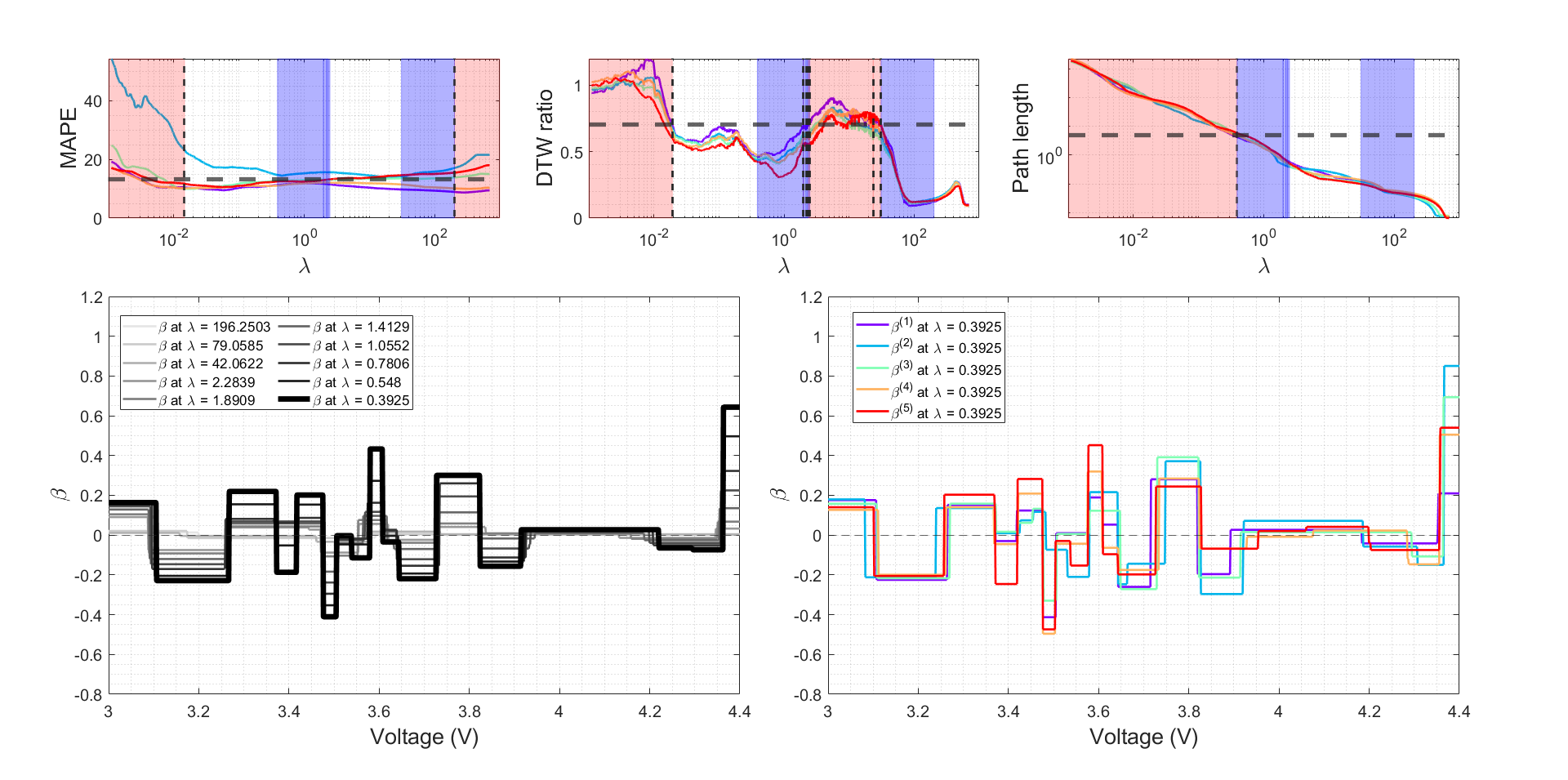}
}
\subfigure[$Q^{\text{B}}(V)$ for Outer loop 3.]{
\includegraphics[width=.470\textwidth]{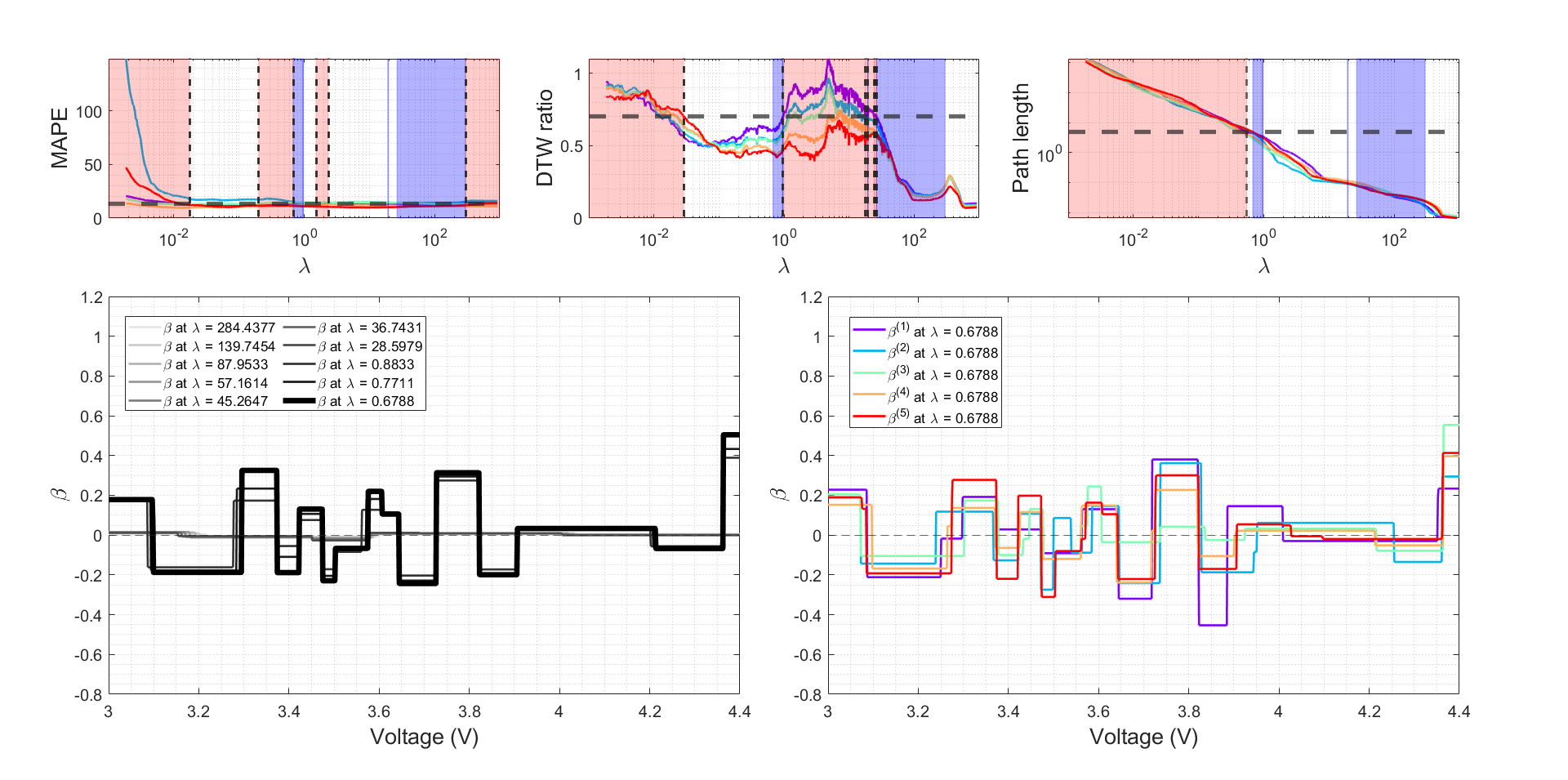}
}
\subfigure[$Q^{\text{B}}(V)$ for Outer loop 4.]{
\includegraphics[width=.470\textwidth]{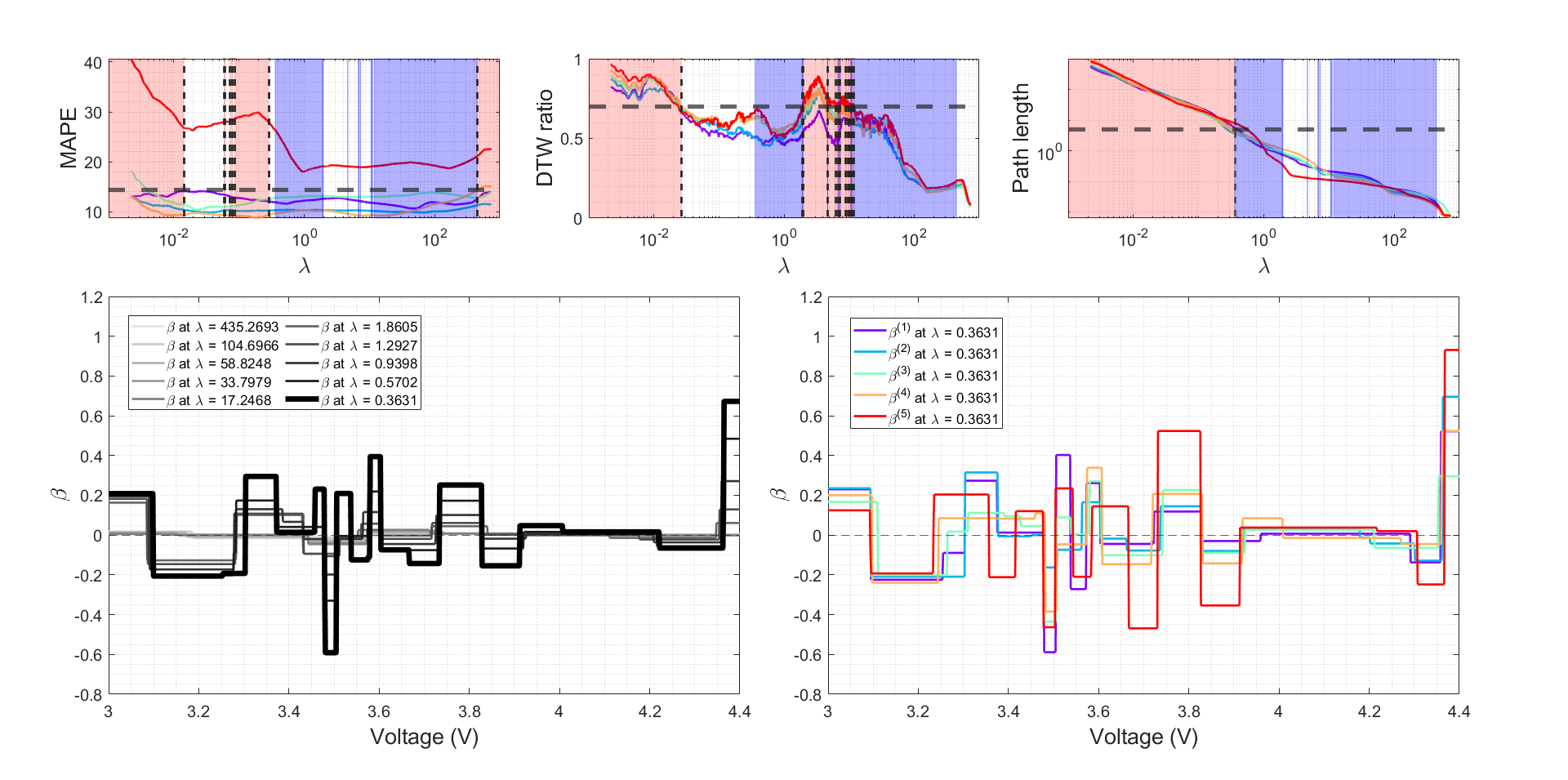}
}
\subfigure[$Q^{\text{B}}(V)$ for Outer loop 5.]{
\includegraphics[width=.470\textwidth]{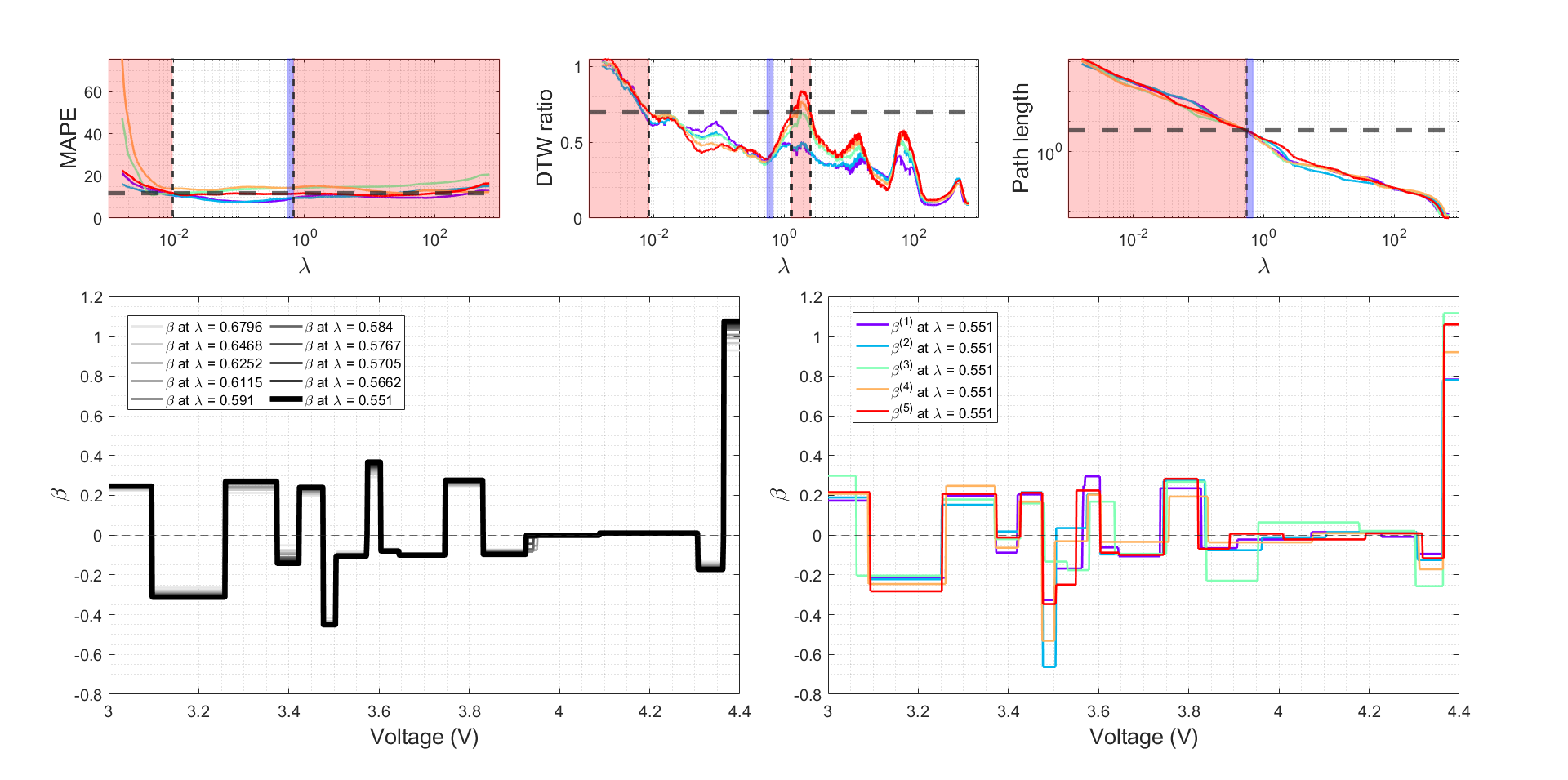}
}
\caption{Three metrics as a function of $\lambda$ when using $Q^{\text{B}}(V)$ as input data for each outer loop.}
\label{fig:appendix_lambda_BQV}
\end{figure}

\newpage
\section{Feature design results with other input data types}
\label{appdx:feature_design_otherinput}
\begin{figure}[htbp!]
\centering
\subfigure[$V^{\text{B}}(\tilde{t})$ for Outer loop 1.]{
\includegraphics[width=.470\textwidth]{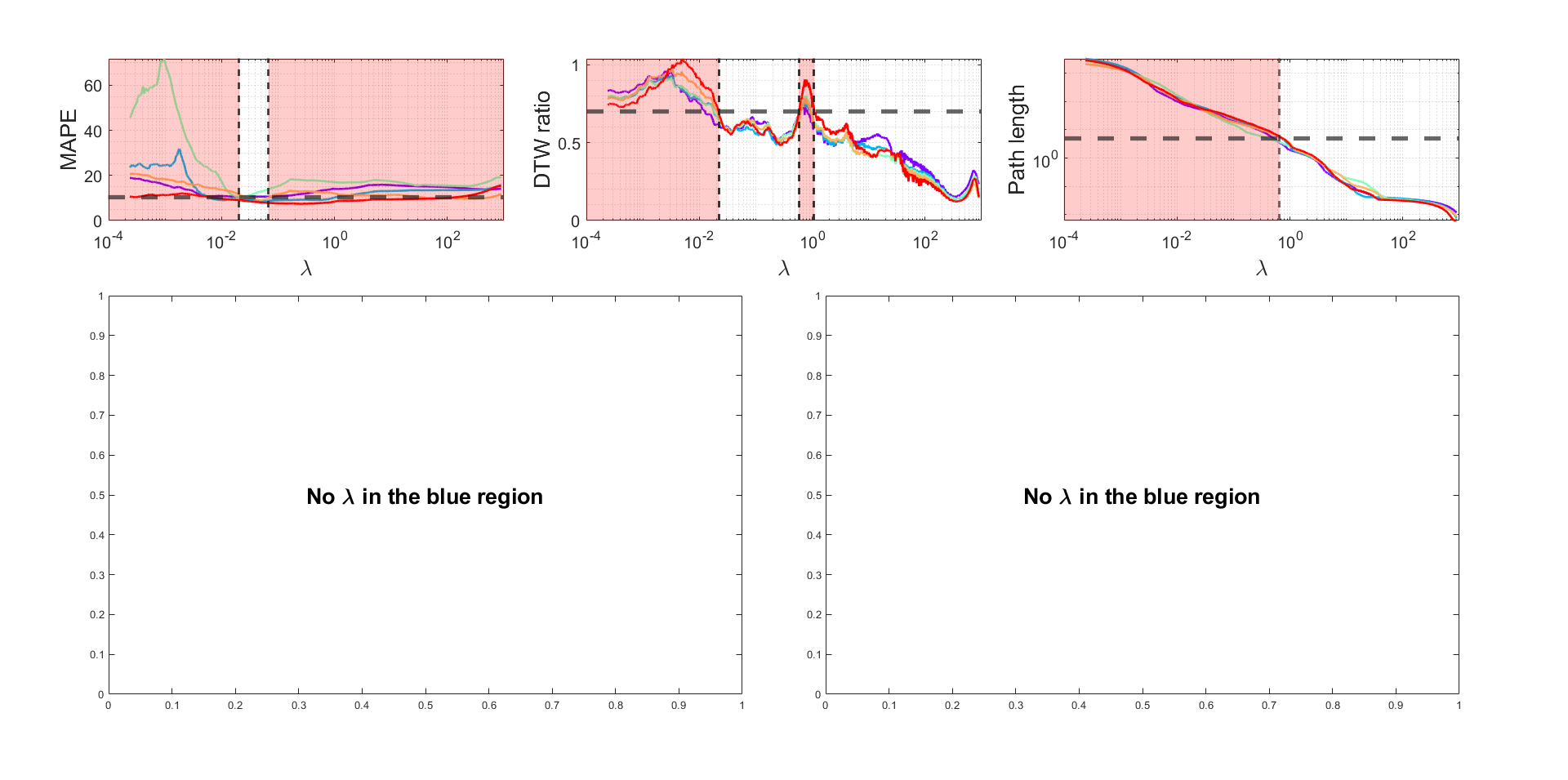}
}
\subfigure[$V^{\text{B}}(\tilde{t})$ for Outer loop 2.]{
\includegraphics[width=.470\textwidth]{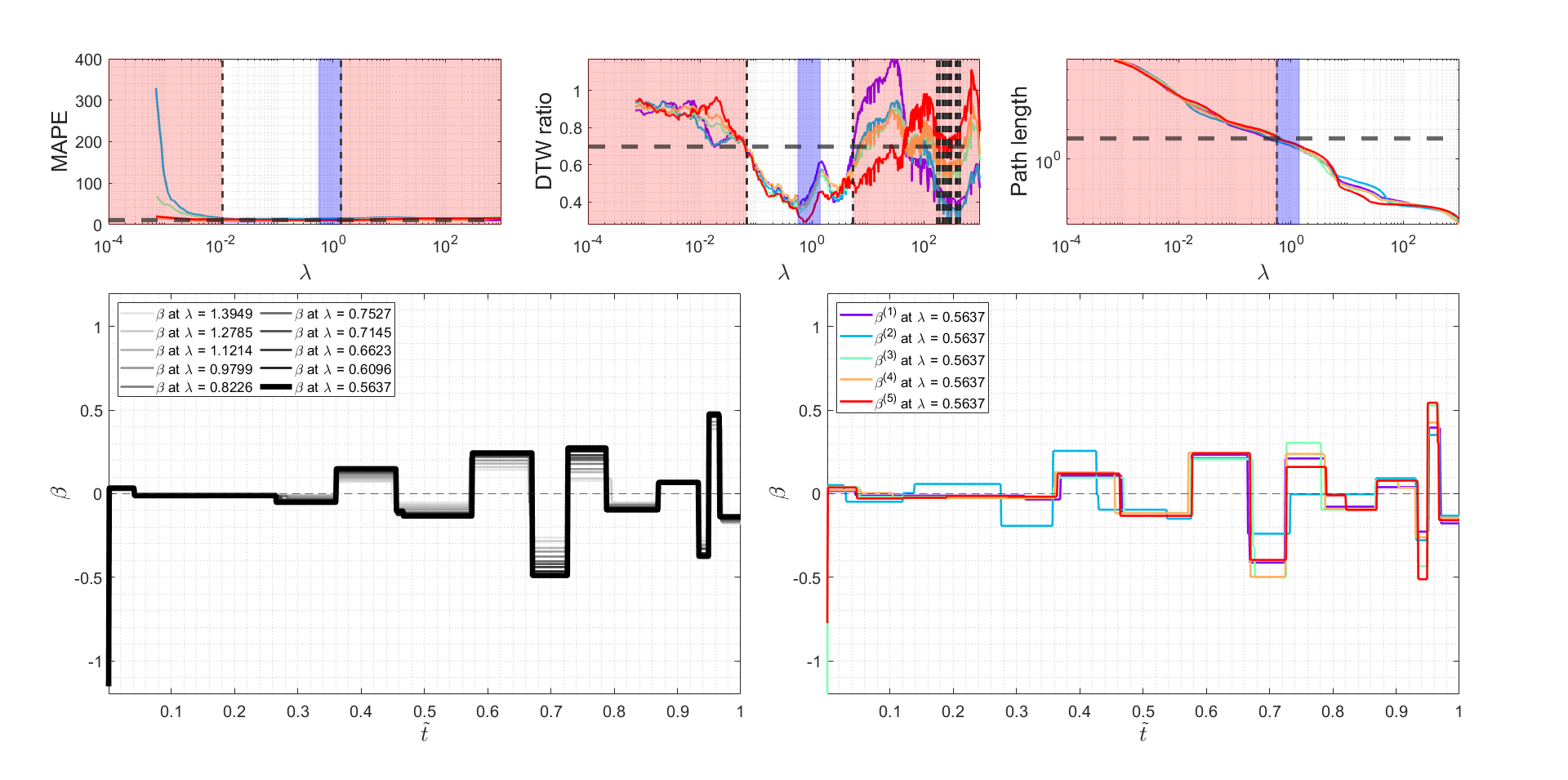}
}
\subfigure[$V^{\text{B}}(\tilde{t})$ for Outer loop 3.]{
\includegraphics[width=.470\textwidth]{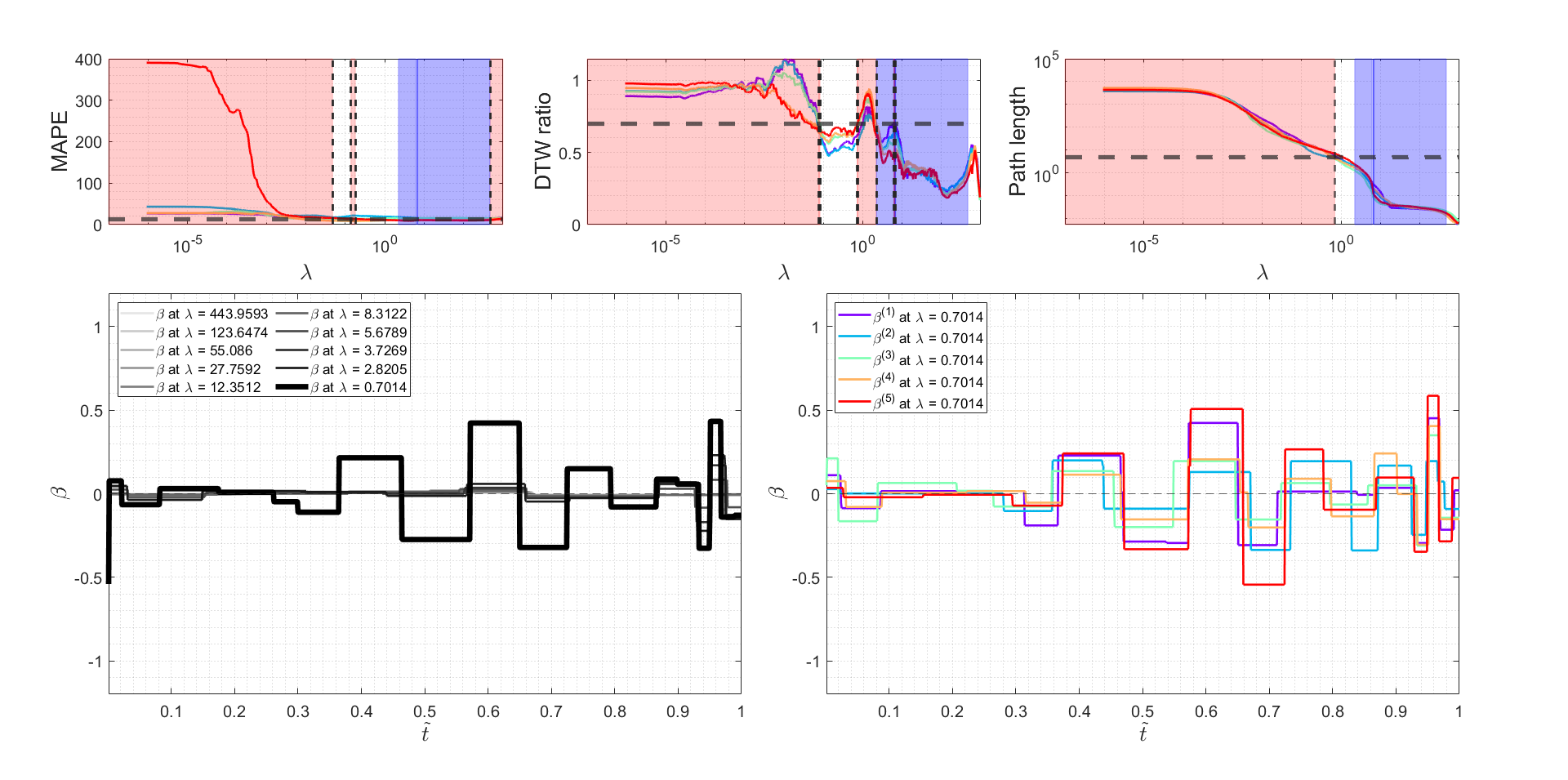}
}
\subfigure[$V^{\text{B}}(\tilde{t})$ for Outer loop 4.]{
\includegraphics[width=.470\textwidth]{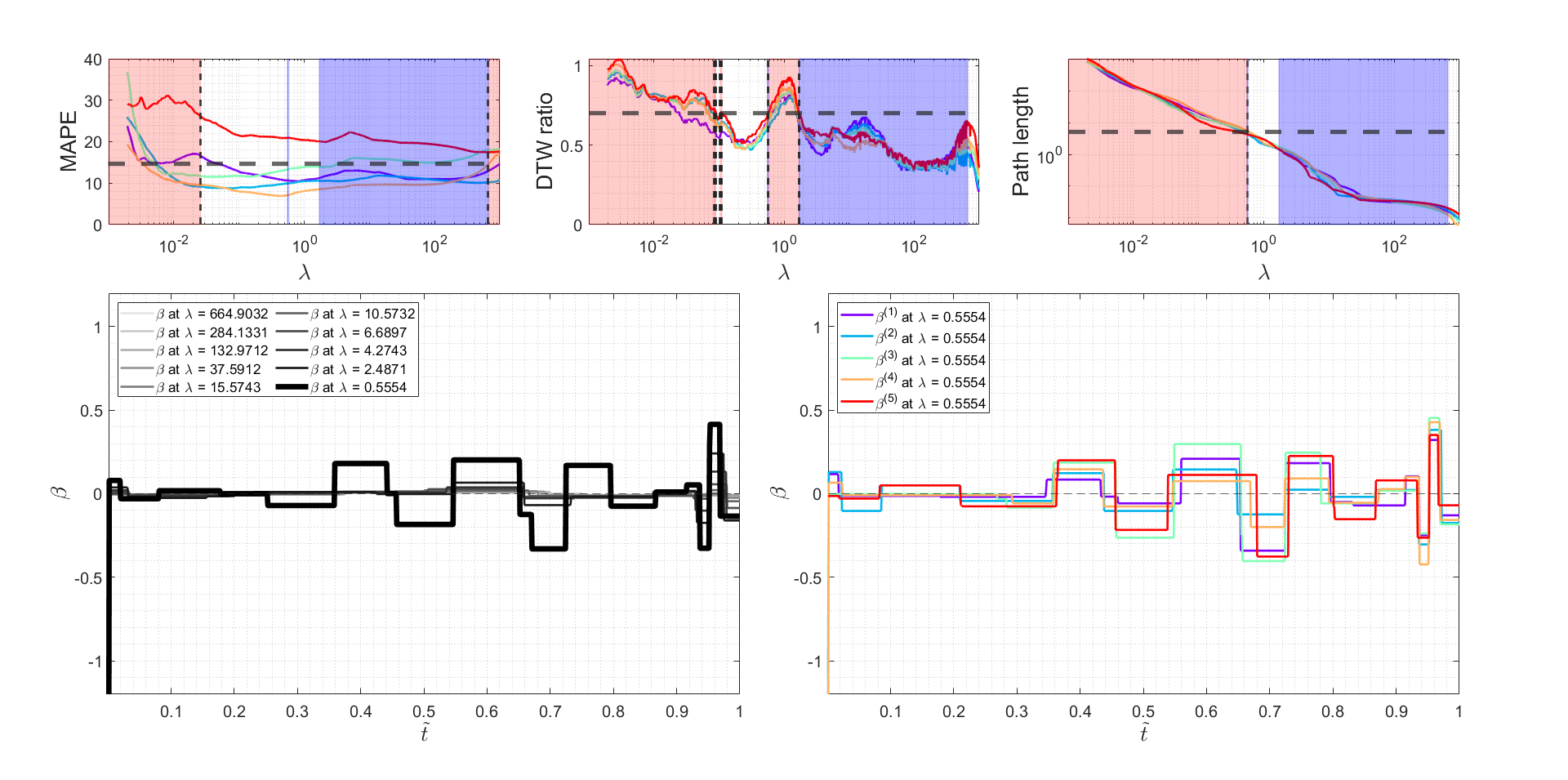}
}
\subfigure[$V^{\text{B}}(\tilde{t})$ for Outer loop 5.]{
\includegraphics[width=.470\textwidth]{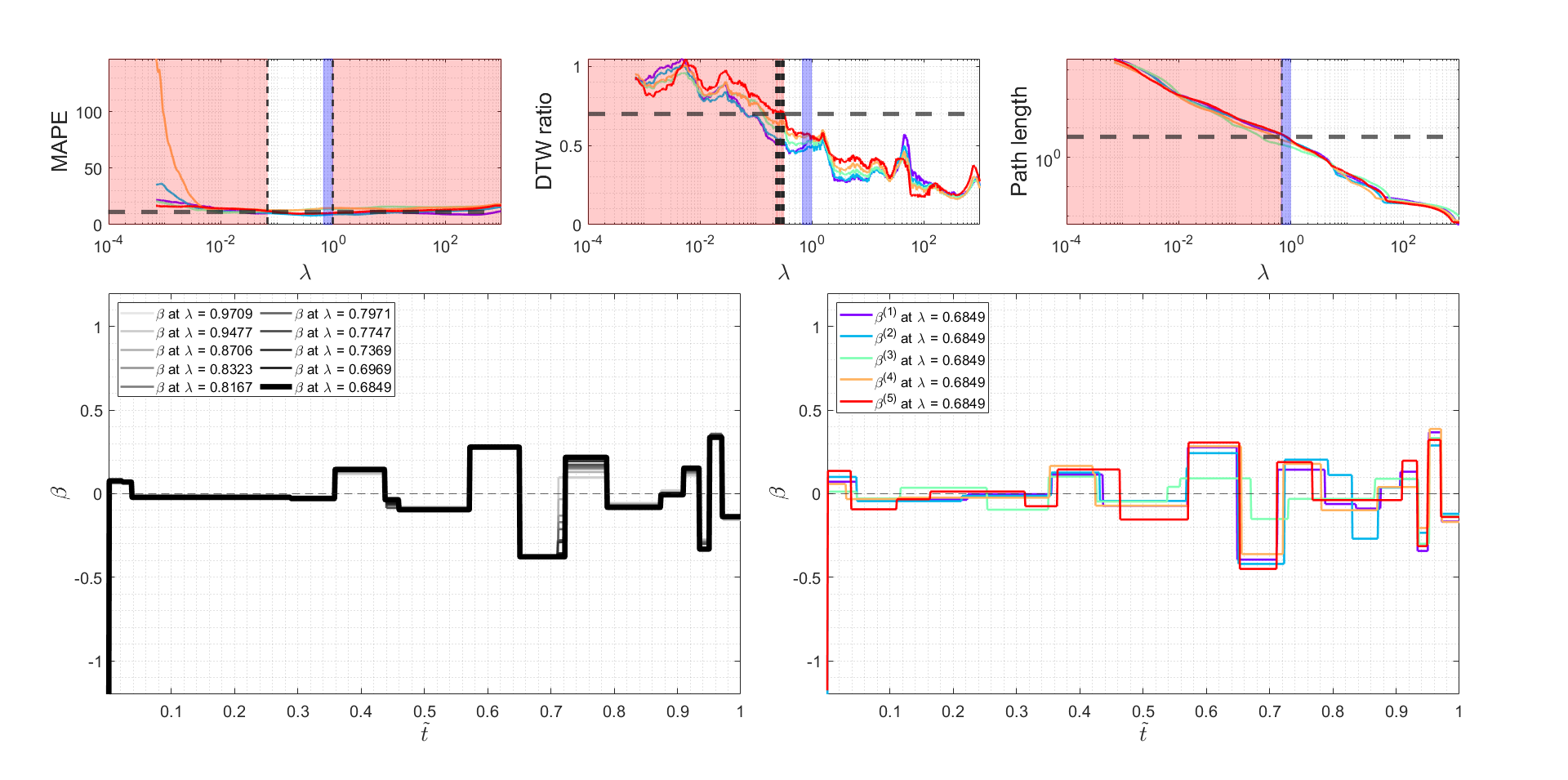}
}
\caption{Three metrics as a function of $\lambda$ when using $V^{\text{B}}(\tilde{t})$ as input data for each outer loop. There is no blue region for Outer loop 1.}
\label{fig:appendix_lambda_BVt}
\end{figure}

\begin{table}[h!]
\renewcommand{\arraystretch}{1.2}
\begin{center}
\begin{tabular}{|c|c|c|}
\hline
Outer & $\lambda$ & Selected features\\ 
 \hline
1 & -- & --\\ \hline
2 & 0.5637 & $V^{\text{B}}(\tilde{t}=0.93)-V^{\text{B}}(\tilde{t}=0.95)$\\ \hline
3 & 0.7014 & \begin{tabular}{@{}c@{}} $V^{\text{B}}(\tilde{t}=0.93)-V^{\text{B}}(\tilde{t}=0.95)$ \\ mean $V^{\text{B}}(\tilde{t}=0.37-0.46)$\end{tabular} \\ \hline
4 & 0.5554 & \begin{tabular}{@{}c@{}}mean $V^{\text{B}}(\tilde{t}=0.97-1)$ \\ $V^{\text{B}}(\tilde{t}=0.55)-V^{\text{B}}(\tilde{t}=0.65)$\end{tabular}\\ \hline
5 & 0.6849 & $V^{\text{B}}(\tilde{t}=0.91)-V^{\text{B}}(\tilde{t}=0.94)$\\ \hline
\end{tabular}
\caption{Feature design results for each outer loop when using $V^{\text{B}}(\tilde{t})$.}
\label{tab:feature_design_BVt}
\end{center}
\end{table}

\begin{figure}[htbp!]
\centering
\subfigure[$Q^{\text{C}}(V)$ for Outer loop 1.]{
\includegraphics[width=.470\textwidth]{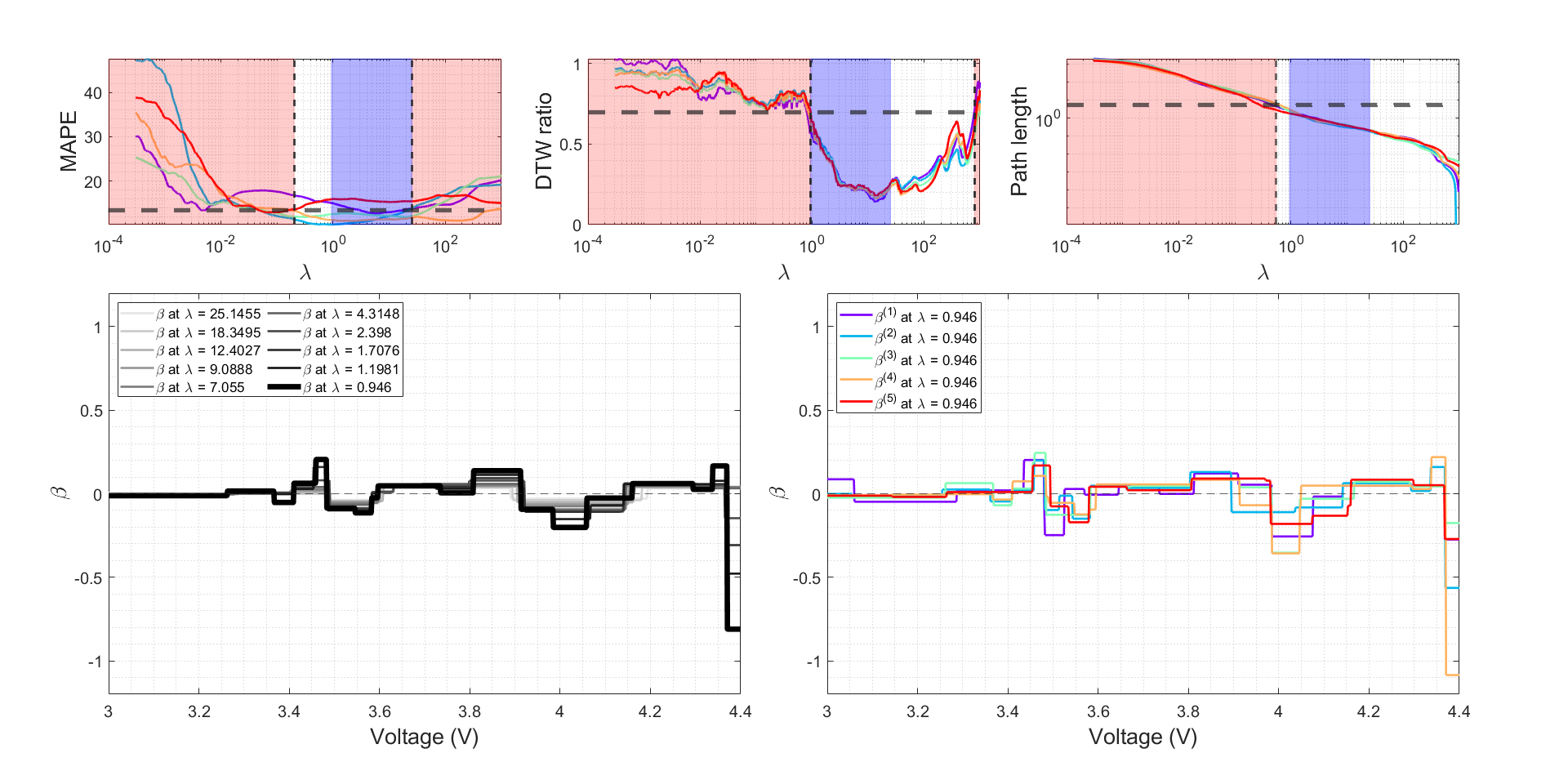}
}
\subfigure[$Q^{\text{C}}(V)$ for Outer loop 2.]{
\includegraphics[width=.470\textwidth]{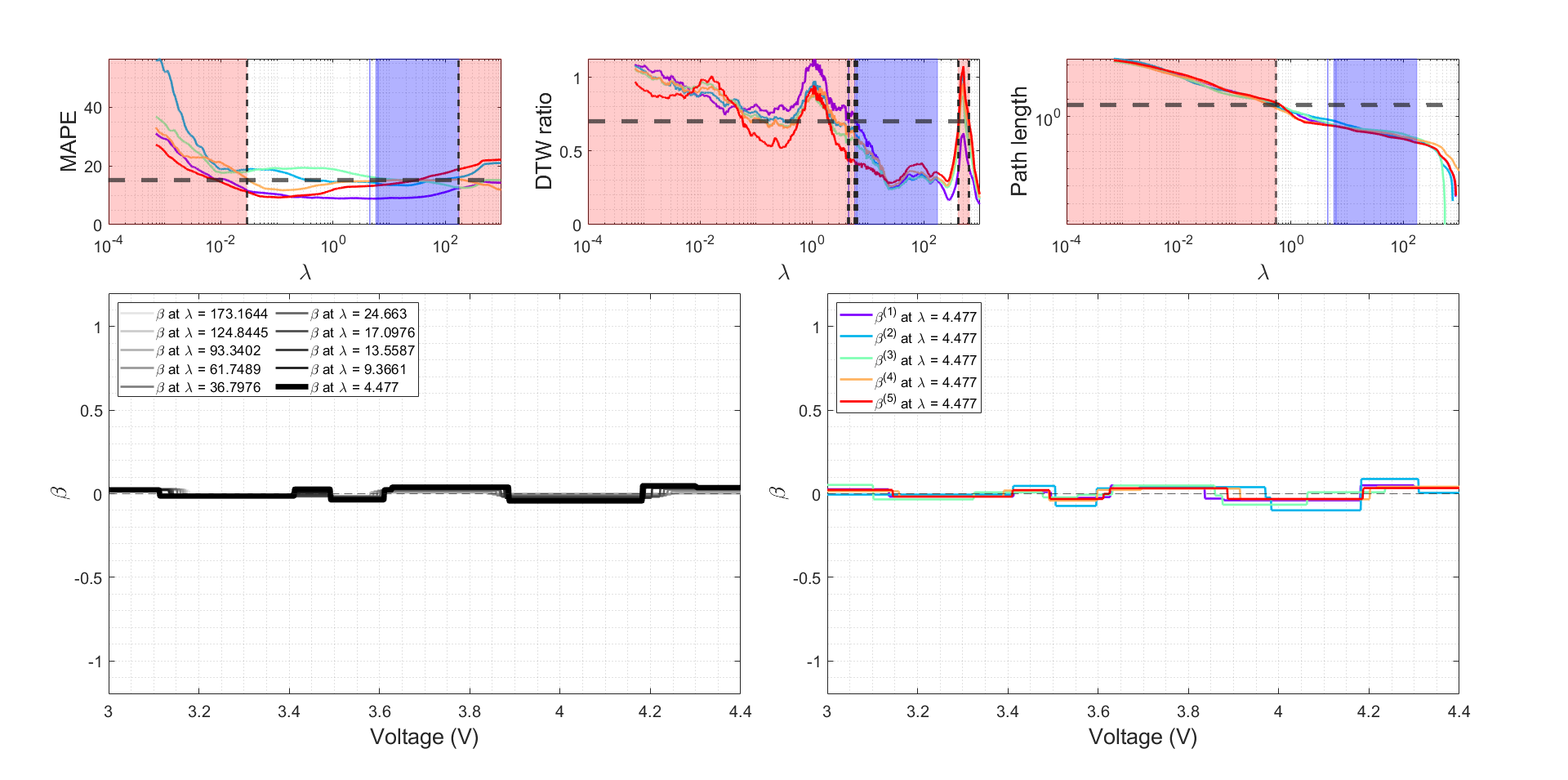}
}
\subfigure[$Q^{\text{C}}(V)$ for Outer loop 3.]{
\includegraphics[width=.470\textwidth]{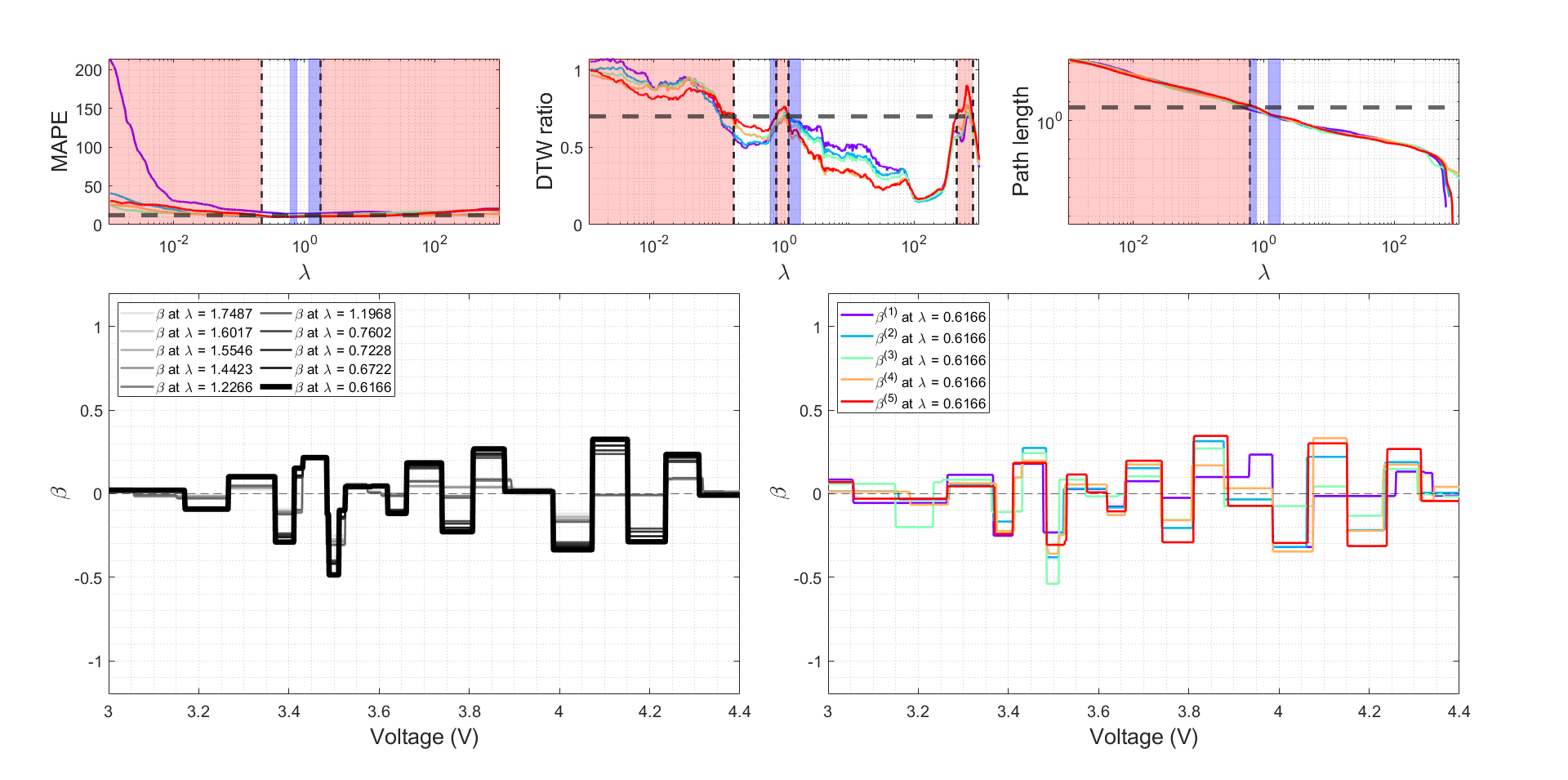}
}
\subfigure[$Q^{\text{C}}(V)$ for Outer loop 4.]{
\includegraphics[width=.470\textwidth]{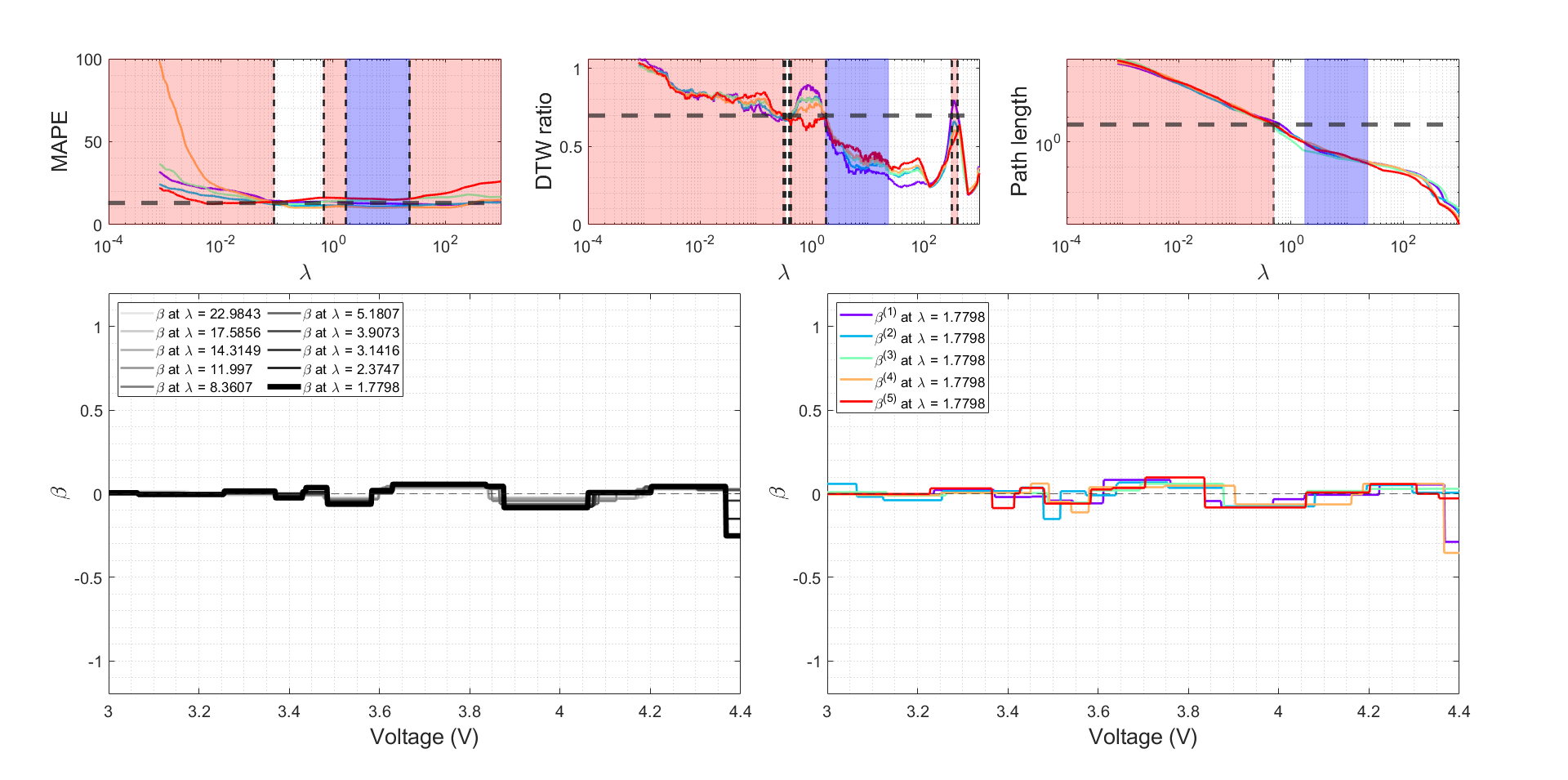}
}
\subfigure[$Q^{\text{C}}(V)$ for Outer loop 5.]{
\includegraphics[width=.470\textwidth]{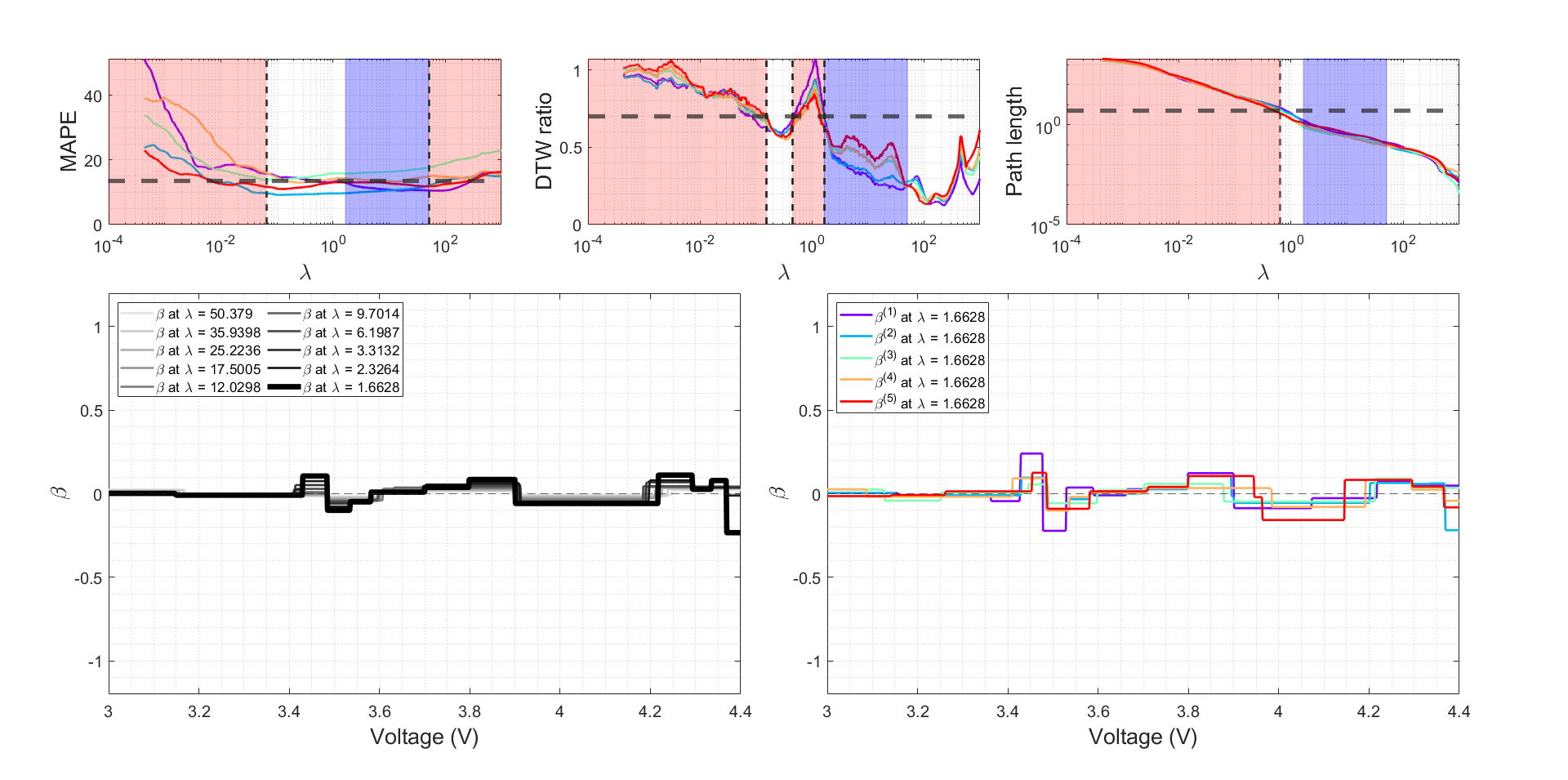}
}
\caption{Three metrics as a function of $\lambda$ when using $Q^{\text{C}}(V)$ as input data for each outer loop.}
\label{fig:appendix_lambda_CQV}
\end{figure}

\begin{table}[h!]
\renewcommand{\arraystretch}{1.2}
\begin{center}
\begin{tabular}{|c|c|c|}
\hline
Outer & $\lambda$ & Selected features\\ 
 \hline
1 & 0.9460 & $Q^{\text{C}}(\SI{3.37}{V})-Q^{\text{C}}(\SI{3.41}{V})$\\ \hline
2 & 4.4770 & \begin{tabular}{@{}c@{}}$Q^{\text{C}}(\SI{3.00}{V})-Q^{\text{C}}(\SI{3.11}{V})$ \\ mean $Q^{\text{C}}(\SI{3.11}{V}-\SI{3.41}{V})$\end{tabular}\\ \hline
3 & 0.6166 & $Q^{\text{C}}(\SI{3.52}{V})-Q^{\text{C}}(\SI{3.62}{V})$ \\ \hline
4 & 1.7798 & $Q^{\text{C}}(\SI{3.58}{V})-Q^{\text{C}}(\SI{3.63}{V})$\\ \hline
5 & 1.6628 & \begin{tabular}{@{}c@{}}$Q^{\text{C}}(\SI{3.00}{V})-Q^{\text{C}}(\SI{3.15}{V})$ \\ mean $Q^{\text{C}}(\SI{3.43}{V}-\SI{3.48}{V})$\end{tabular}\\ \hline
\end{tabular}
\caption{Feature design results for each outer loop when using $Q^{\text{C}}(V)$.}
\label{tab:feature_design_CQV}
\end{center}
\end{table}

\begin{figure}[htbp!]
\centering
\subfigure[$V^{\text{C}}(\tilde{t})$ for Outer loop 1.]{
\includegraphics[width=.470\textwidth]{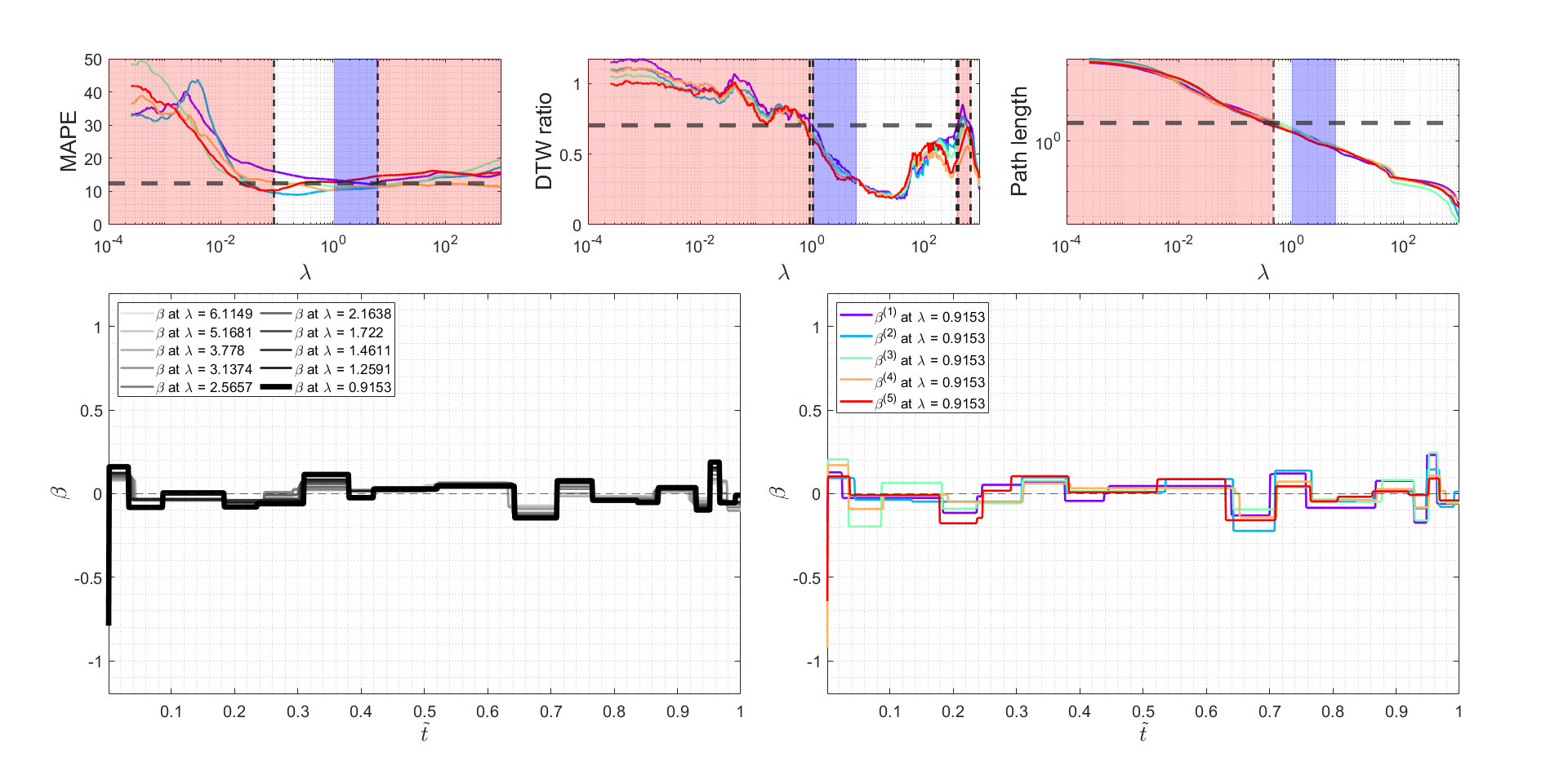}
}
\subfigure[$V^{\text{C}}(\tilde{t})$ for Outer loop 2.]{
\includegraphics[width=.470\textwidth]{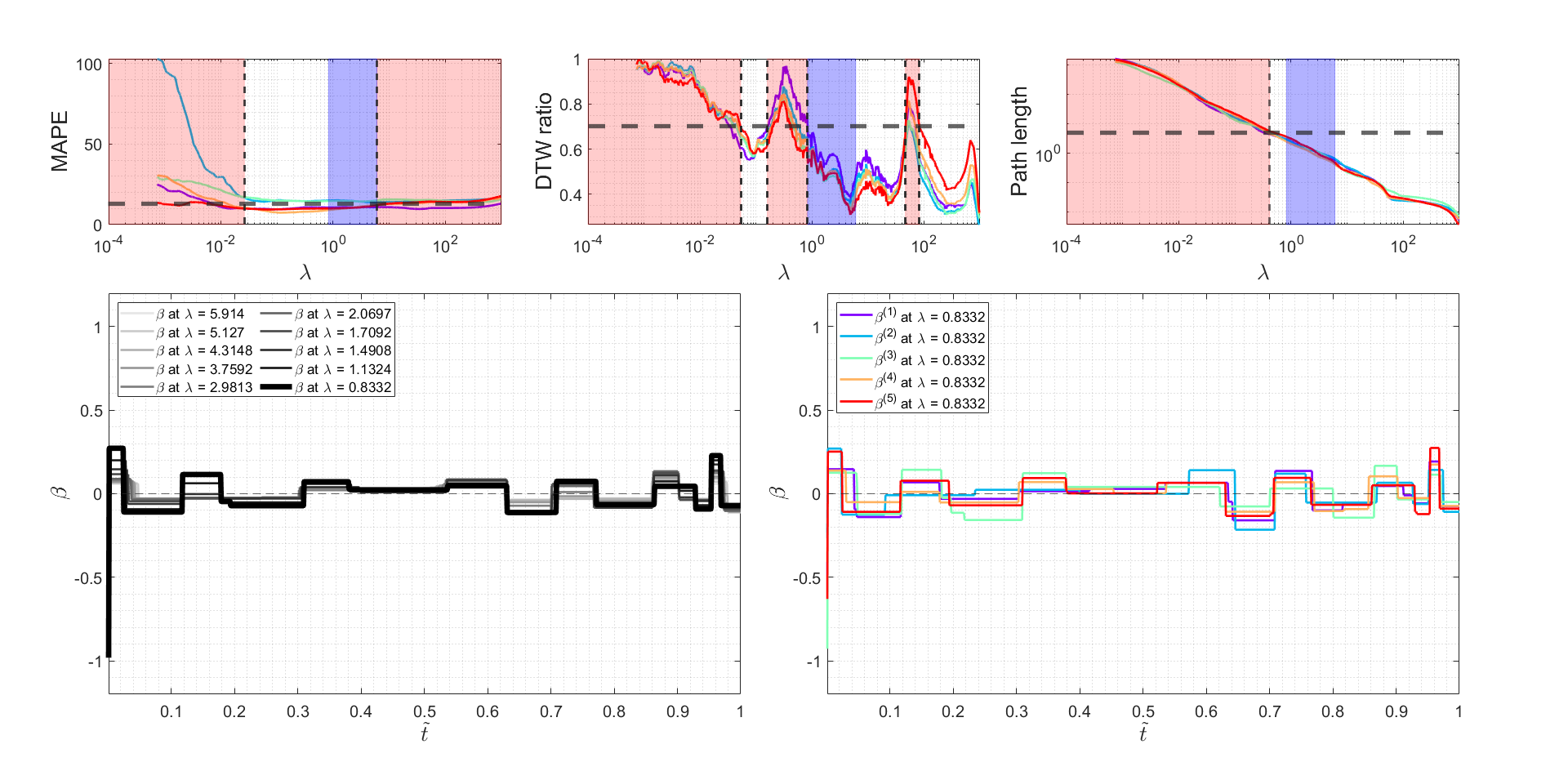}
}
\subfigure[$V^{\text{C}}(\tilde{t})$ for Outer loop 3.]{
\includegraphics[width=.470\textwidth]{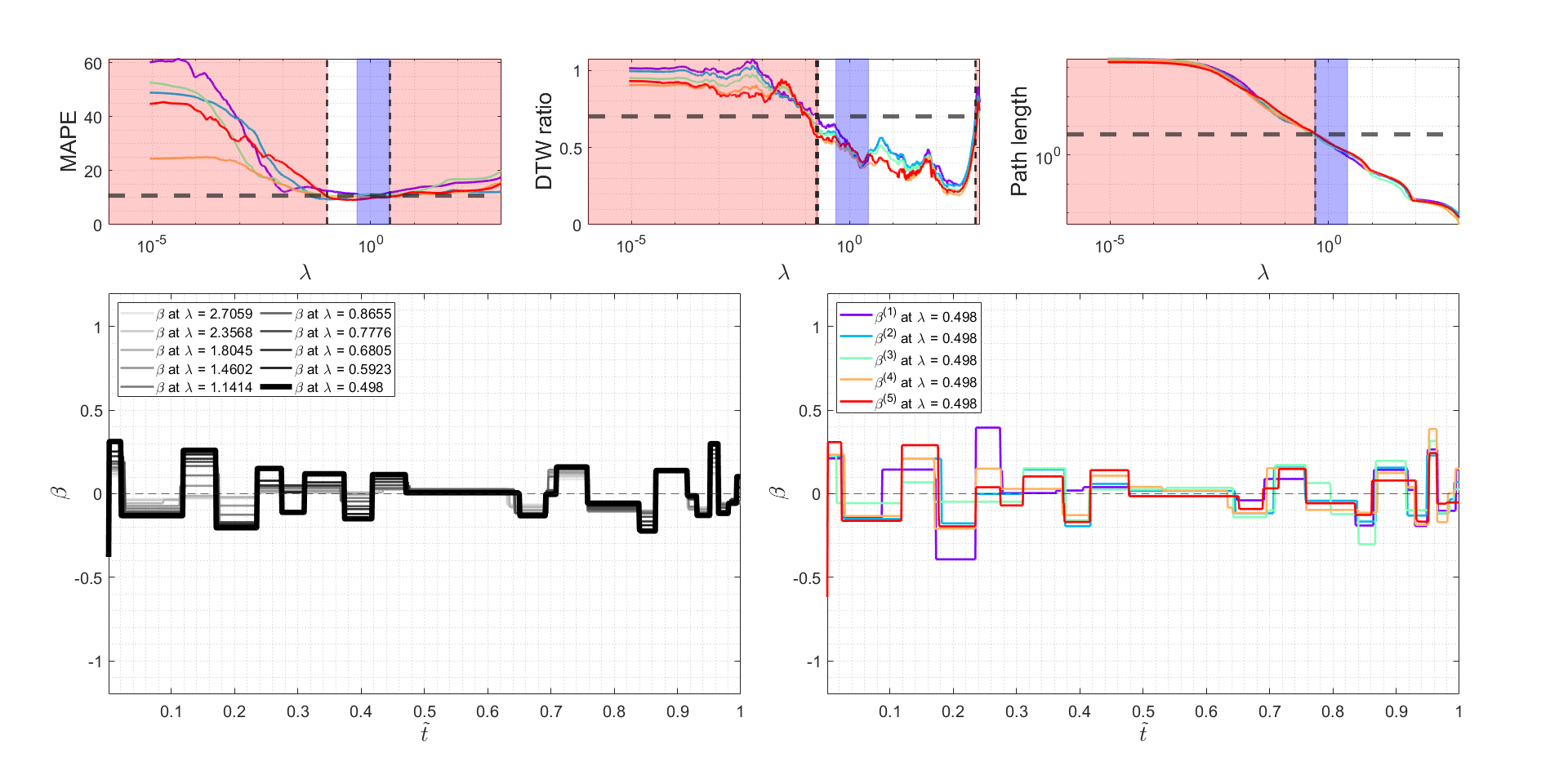}
}
\subfigure[$V^{\text{C}}(\tilde{t})$ for Outer loop 4.]{
\includegraphics[width=.470\textwidth]{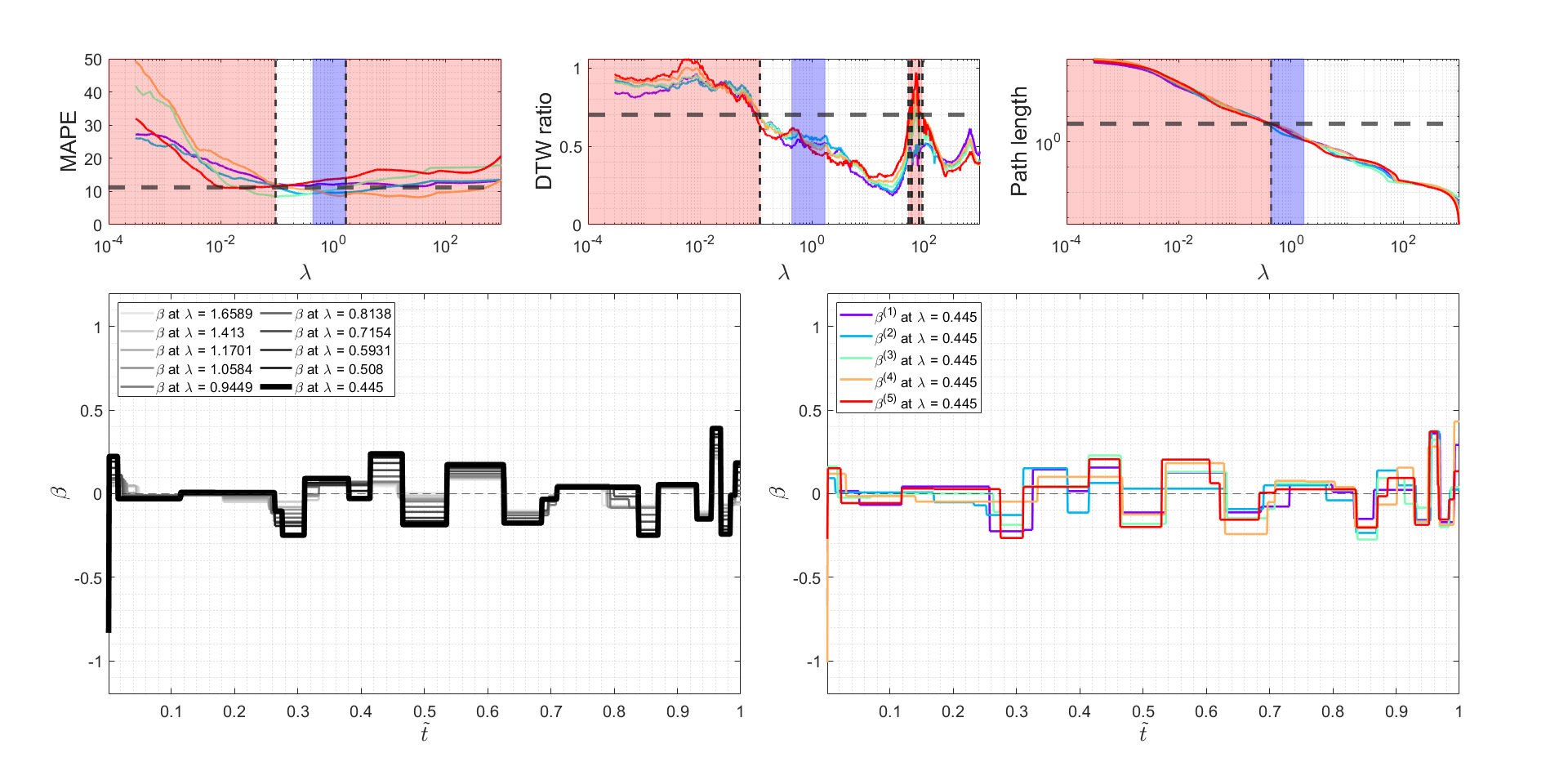}
}
\subfigure[$V^{\text{C}}(\tilde{t})$ for Outer loop 5.]{
\includegraphics[width=.470\textwidth]{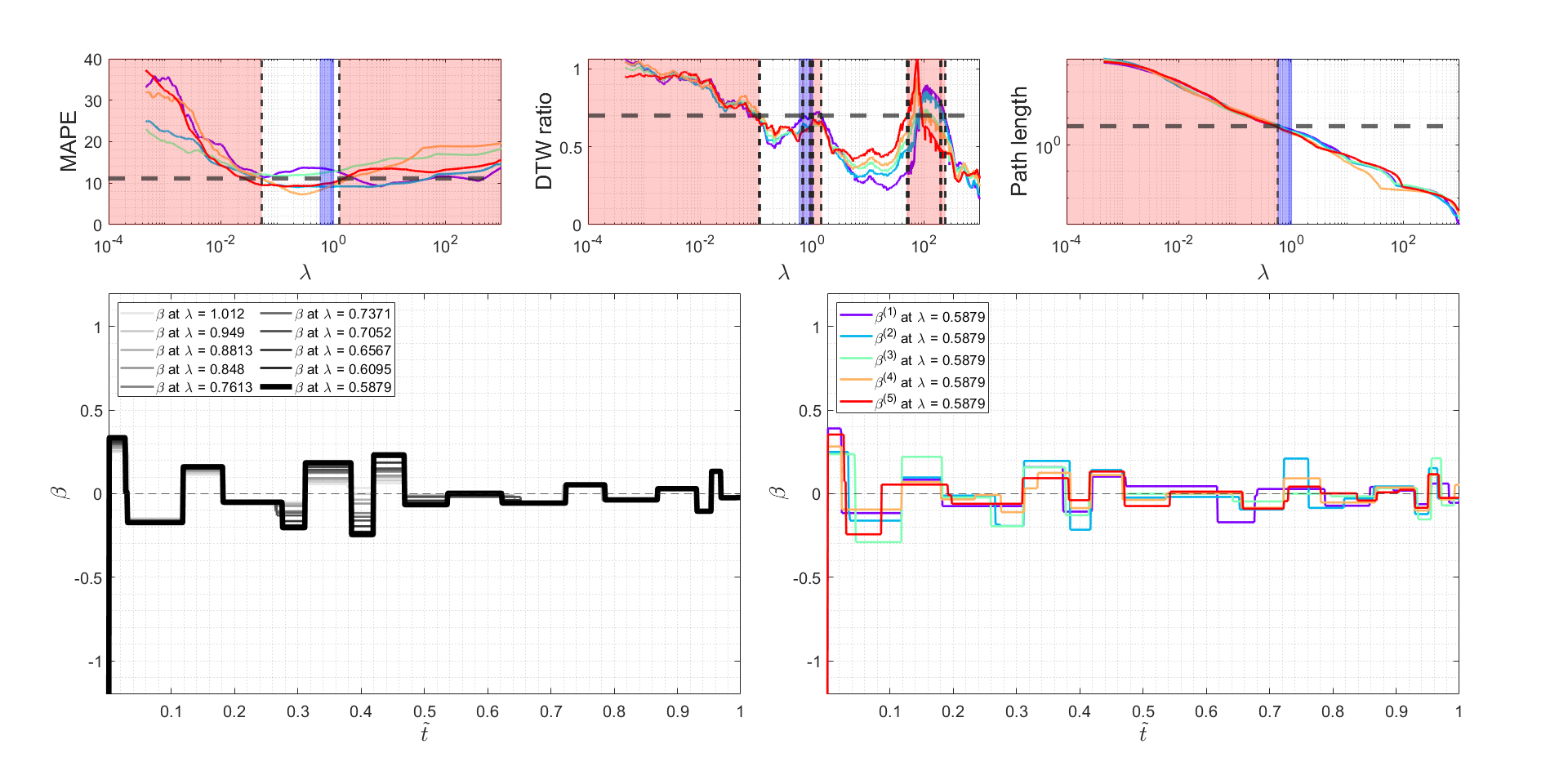}
}
\caption{Three metrics as a function of $\lambda$ when using $V^{\text{C}}(\tilde{t})$ as input data for each outer loop.}
\label{fig:appendix_lambda_CVt}
\end{figure}

\begin{table}[h!]
\renewcommand{\arraystretch}{1.2}
\begin{center}
\begin{tabular}{|c|c|c|}
\hline
Outer & $\lambda$ & Selected features\\ 
 \hline
1 & 0.9153 & mean $V^{\text{C}}(\tilde{t}=0.99-1)$\\ \hline
2 & 0.8332 & mean $V^{\text{C}}(\tilde{t}=0.97-1)$\\ \hline
3 & 0.4980 & $V^{\text{C}}(\tilde{t}=0.99)-V^{\text{C}}(\tilde{t}=1)$ \\ \hline
4 & 0.4450 & mean $V^{\text{C}}(\tilde{t}=0.99-1)$\\ \hline
5 & 0.5879 & mean $V^{\text{C}}(\tilde{t}=0.93-0.95)$\\ \hline
\end{tabular}
\caption{Feature design results for each outer loop when using $V^{\text{C}}(\tilde{t})$.}
\label{tab:feature_design_CVt}
\end{center}
\end{table}

\newpage
\section{Physics-Based Models}

\subsection{Electrode Chemical Potential Fitting \label{sec:fit_OCV}}

\begin{figure}[h!]
\includegraphics{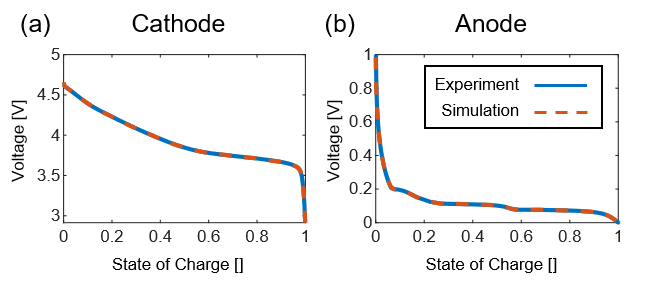}
\centering
\vspace{-0.3cm}
\caption{Comparison of simulated open circuit voltage and experimental half-cell measurements for cathode (a) and anode (b).}
\label{fig:OCV_fits}
\end{figure}

The chemical potential as a function of the State-of-Charge (SoC) for both the cathode and anode active material is critical in the reactive particle ensemble model used in this study. An approximation of the open circuit voltage (OCV) is obtained from half cells cycled at C/20. In these experiments, we have assumed that the reference electrode is not kinetically limiting, so the OCV approximates the working electrode's chemical potential by relationship of $V_{\text{OCV}} = - e \mu$. In this work, following Zhao et al.~\cite{zhao_learning_2020,zhao_learning_2023}, the chemical potential is decomposed into (1) entropic component based on a lattice model and (2) enthalpic component defined by a Legendre polynomial series expansion as
\begin{gather}
    V_{\text{OCV}}(c) = - \frac{k_\text{B} T_{\text{ref}}}{e} \ln\!{\left( \frac{c}{1-c} \right)} + \sum_{i=0}^N a_n P_n([2c-1]).
\end{gather}
The coefficients for the polynomial series are learned through linear regression. The OCV fittings for the cathode and anode systems are shown in Figure \ref{fig:OCV_fits} and the values for the coefficients are in Table \ref{tab:OCV_fits}.

\begin{table}[h!]
\begin{center}
\small 
\begin{tabular}{|c|c|c|}
\hline
$a_n$ & Cathode - NMC 532 & Anode - Artificial Graphite \\ \hline
0 & 3.9441 & 0.1177 \\
1 & $-0.4024$ & $-0.0352$ \\
2 &  $0.1444$ & $0.0801$ \\
3 & $-0.0516$ & $-0.0664$ \\
4 & $-0.0735$ & 0.0713 \\
5 & $-0.0541$ & $-0.0662$ \\
6 & $-0.0405$ & 0.0507 \\
7 & $-0.0437$ & $-0.0427$ \\
8 & $-0.0627$ & 0.0543 \\
9 & $-0.0442$ & $-0.0440$ \\
10 & $-0.0499$ & 0.0294 \\
11 & $-0.0375$ & $-0.0099$ \\
12 & $-0.0416$ & 0.0118 \\
13 & $-0.0353$ & $-0.0014$ \\
14 & $-0.0279$ & $-0.0005$ \\
15 & $-0.0255$ & $-0.0001$ \\
16 & $-0.0214$ & 0.0101 \\
17 & $-0.0165$ & $-0.0071$ \\
18 & $-0.0176$ & 0.0064 \\
19 & $-0.0124$ & $-0.0087$ \\
20 & & 0.0111 \\
21 & & $-0.0050$ \\
22 & & 0.0009 \\
23 & & 0.0002 \\
24 & & $-0.0007$ \\ \hline

\end{tabular}
\caption{Coefficients for Legendre Polynomials used in OCV fitting.}
\label{tab:OCV_fits}
\end{center}
\end{table}

\subsection{Electrode Utilization Model \label{sec:fit_utilization}}

\begin{figure}[h!]
\includegraphics[width=0.75\textwidth]{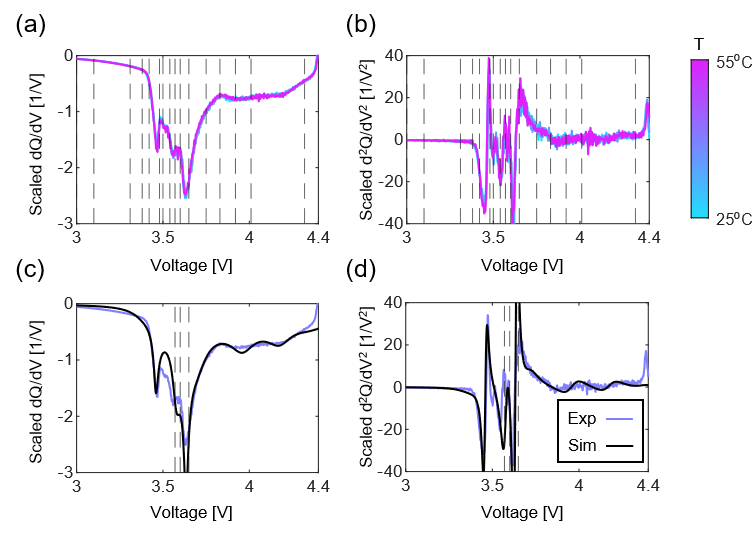}
\centering
\vspace{-0.3cm}
\caption{(a) Differential capacitance as a function of cell voltage for a post-formation C/20 rate performance test for a subset of `extreme slow formation' cells across different temperatures. (b) d$^2Q$/d$V^2$ for same cells. (c) Simulated differential capacitance from utilization model for average `extreme slow formation' cells electrochemical features. (d) d$^2Q$/d$V^2$ for same simulation.  Designed voltage features in Table 6 shown as vertical lines to show connections to plotted electrochemical features.}
\label{fig:utilization_fits}
\end{figure}

A model for the electrode utilization is created in the limit of no transport and kinetic limitations.
This can happen if the battery process time scales are small, but is also the case where the operation timescale is long.
We apply this simplified model to the study of a C/20 set of rate performance taken post-formation.
In this model, the system is parameterized by four parameters: fraction of cathode capacity `active' to filling/emptying $\beta_\text{c}$, fraction of anode capacity `active' to filling/emptying $\beta_\text{a}$, remaining lithium inventory capacity $Q_\text{rem}$, and voltage shift due to external resistances $V_{\text{shift}}$.
These parameters consist the system's utilization state resulting from formation.
In the limits described above, the relationship between the total current and the voltage are described by
\begin{gather}
    \frac{I}{Q_{\text{c,total}}} = - \frac{Q_{\text{a,total}}}{Q_{\text{c,total}}}  \beta_\text{a} \frac{\partial \bar{c}_\text{a}}{\partial t} = \beta_\text{c}\frac{\partial \bar{c}_\text{c}}{\partial t}, \\
    V_{\text{cell}} = - \frac{1}{e} (\mu_\text{c} (\bar{c}_\text{c}) - \mu_\text{a} (\bar{c}_\text{a})) - V_{\text{shift}},
\end{gather}
where $\bar{c}_j$ is the average normalized concentration of the species in the electrode.
The ratio $Q_{\text{a,total}}$/$Q_{\text{c,total}}$ in this model is equivalent to the N/P ratio of the battery, which is known to be 1.16 for the cells investigated in this work \cite{cui_data-driven_2024}.
The cell voltage dynamics in this model is the difference in the chemical potential functions fitted in Section \ref{sec:fit_OCV} with an offset due to external resistances.
We have focused the analysis of the electrode utilization parameters to a subset of the cells, denoted \gls{esf} cells.
As shown in Figure \ref{fig:utilization_fits}, the differential capacitance feature for the C/20 \gls{rpt} measurements is nearly identical across \gls{esf} cells.
We hypothesize that this observation can be explained by a similar utilization state as a result of formation at slower rates.
This hypothesis can be understood of a self-limiting \gls{sei} layer, where a maximum \gls{sei} capacity can be reached below a critical formation time. 
The average electrode utilization state for these cells is located in Table \ref{tab:utilization_state}.
Interestingly, the learned utilization state indicates that the effective capacity lost at each electrode is greater than the lithium inventory lost, an indication that further understanding needs to be made into both the effective active capacity state variable $\beta_j$ and designing batteries where these values are maximized.
As seen in Figure \ref{fig:utilization_fits}, the electrode utilization captures the broad features of the experimental dataset, but still fails to fit the differential capacitance versus voltage curve perfectly.
This is likely due to the failure of the assumption that the process timescale is large enough to neglect the reaction and transport limitations of the system is inaccurate. 

\begin{table}[h!]
\begin{center}
\begin{tabular}{|c|c|c|}
\hline
State Variable & Value & Units \\ \hline
$\beta_\text{c}$ & 0.911 & - \\
$\beta_\text{a}$ & 0.854 & - \\
$\frac{Q_\text{rem}}{Q_{\text{c,total}}}$ & 0.930 & - \\
$V_{\text{shift}}$ & 0.014 & V \\ \hline
\end{tabular}
\caption{Learned utilization state for \gls{esf} cells.}
\label{tab:utilization_state}
\end{center}
\end{table}

\subsection{\label{si-model}Reactive Particle Ensemble Model}
The reactive particle ensemble model assumes that each electrode can be approximated by a set of $N_\text{particles}$ each of which is approximated as an electrochemically reactive surface.
The particles are nearly indistinguishable, except for the surface rate constant pre-factor, $k_{0}$, which is defined separately for each particle.
The state of the $i$th particle in electrode $j$ can be defined solely by the average filling fraction of the particle, $\bar{c}^{(i)}_j$, where $j$ can either be the cathode `c' or anode `a', whereas $i$ can take on any positive integer value below the number of simulated microscopic particles, $N_j$. The evolution equations for each particle are 
\begin{gather}
    \beta_j \frac{\partial \bar{c}^{(i)}_j}{\partial t} = \frac{a_{V,j}}{e} j (\bar{c}^{(i)}_j, V_j, T), \\
    j(\bar{c}^{(i)}_j, V_j, T) = k_{0,j}^{(i)} \exp\!{\left(\frac{-E_\text{A}}{R} \!\left(\frac{1}{T} - \frac{1}{T_\text{ref}} \right)\right)} (1-\bar{c}^{(i)}_j) \sqrt{\bar{c}^{(i)}_j} \sinh\!{ \left(- \frac{e \eta(\bar{c}^{(i)}_j, V_j)}{k_\text{B} T}\right)}, \label{eq:ICET} \\
    \eta(\bar{c}^{(i)}_j, V_j) = \frac{1}{e} \mu_j (\bar{c}^{(i)}) + V_j.
\end{gather}
The rate expression for the Faradaic current density, Equation \ref{eq:ICET}, is predicted by the quantum theory of ion-coupled electron transfer (\gls{icet})~\cite{bazant_unified_2023}. The \gls{icet} rate expression has a symmetric Butler-Volmer dependence on overpotential (for equal oxidation and reduction ion transfer energies), but an asymmetric dependence on filling fraction for intercalation reactions, which has been verified by learning from X-ray images of nanoparticles~\cite{zhao_learning_2023}.
The chemical potential for the cathode and anode particles is taken as $\mu_j (\bar{c}) = - V_{\text{OCV}, j} (\bar{c}) / e $ where the open circuit voltage fits are shown in Section $\ref{sec:fit_OCV}$.
For any given particle in this ensemble model approach, the resistance is the specific charge transfer resistance, $R_{\text{ct}}^{(i)} = \frac{k_\text{B} T }{e k_0^{(i)}}$.
One way to conceptualize this model is as a resistance distribution model, but where non-linear dynamics are considered solely from the reaction kinetics.
The connection between particle-level microscopic current densities and macroscopic current is
\begin{gather}
    \frac{I}{Q_{\text{c,total}}} = \frac{\text{C-rate}}{3600 \text{s}} = -\!\left(\frac{\nu_\text{a} a_{\text{V,a}}}{e c_{\text{ref,c}} \nu_\text{c}}\right)\! \frac{1}{\beta_\text{a} N_\text{a}} \sum_{i=1}^{N_{\text{a}}} j (\bar{c}^{(i)}_\text{a}, V_\text{a}, T) = \left(\frac{a_{\text{V,c}}}{e c_{\text{ref,c}}}\right)\! \frac{1}{\beta_\text{c} N_\text{c}} \sum_{i=1}^{N_{\text{c}}} j (\bar{c}^{(i)}_\text{c}, V_\text{c}, T),
\end{gather}
where the relationship between the macroscopic and microscopic quantities are captured by the area to volume ratio $a_{\text{V}}$ and volume of each electrode $\nu$.
We have simplified these equations in the code implementation by assuming that $(\nu_\text{a} a_{\text{V,a}})/(e c_{\text{ref,c}} \nu_\text{c}) = 1$ and $a_{\text{V,c}} / (e c_{\text{ref,c}}) = 1$.
Though this is likely not true, they are likely to be of the same order of magnitude, if the electrodes are well-designed as the total surface area for reaction should be balanced between cathode and anode.
Additionally, they are simple scaling factors, so the $k_0$ distribution is defined in reference to these scaling factors.
The code is written using forward-Euler methods in time to solve the voltage, particle level \gls{soc}, and total electrode \gls{soc} under a C/5 current constraint. Because each particle in an electrode is assumed to be identical, the relationship between particle and electrode \gls{soc} is simply
\begin{gather}
    \bar{c}_j = \frac{1}{N_j} \sum_{i=1}^{N_{j}} \bar{c}_j^{(i)}.
\end{gather}

The simulation is initialized by using the electrode utilization parameters, as described in \ref{sec:fit_utilization}. Given that dataset `B' always has the same starting cell voltage of 4.4 V, each electrode's \gls{soc} and electrostatic potential $V_j$ can be simulated.
We assume that all battery particles start at the same \gls{soc} at the start ($\bar{c}_j^{(i)}(t=0) = \bar{c}_j (t=0)$).
This is likely a poor assumption given that dataset `B' is not taken after a voltage or OCV hold where the system is given time to equilibrate.
As we can neither quantify nor verify the particle SoC distribution from the experimental dataset, we leave those investigations for future works.
From this model, we are able to simulate electrochemical C/5 discharge, where the output is $V(t)$ and $Q(t)$.
Because this model simplifies the full porous electrode system to a reactive particle ensemble model, the model is specified by a low quantity of free parameters.
Besides the activation energy $E_\text{A}$ which is assumed to be the same for both cathode and anode, the only free parameters are descriptors for defining the distribution of $k_{0}$ being sampled for the electrode.
For the investigations of the cells under \gls{esf} conditions, we define these distributions as single basis log-Gaussians where
\begin{gather}
    \ln {k_{0,j}^{(i)}} \sim \mathcal{N}(\ln{\bar{k}_{0,j}}, \sigma_j).
\end{gather}

For the results shown in Figure 6, we have parameterized the model with $E_\text{A} = 45$ kJ/mole, $\bar{k}_{0,\text{c}} = 5$$\times$$10^{-7}$, $\bar{k}_{0,\text{a}} = 10^{-7}$, $\sigma_\text{c} = 1$, and $\sigma_\text{a} = 0.5$.

\end{document}